\documentclass[11pt]{article}

\usepackage[preprint]{acl}

\usepackage{times}
\usepackage{latexsym}

\usepackage[T1]{fontenc}

\usepackage[utf8]{inputenc}

\usepackage{microtype}

\usepackage{inconsolata}

\usepackage{graphicx}
\usepackage{hyperref}
\usepackage{booktabs}
\usepackage{multirow}
\usepackage{makecell}
\usepackage{subcaption}
\usepackage{etoc}
\usepackage{fancyvrb}
\usepackage{url}
\usepackage{algorithm}
\usepackage{algpseudocode}
\usepackage{wrapfig} 
\usepackage{colortbl}
\usepackage{pifont}
\usepackage[most]{tcolorbox}
\usepackage{xspace}
\usepackage{amsmath}
\usepackage{amssymb}
\usepackage{enumitem}
\usepackage{xcolor}
\usepackage{color}
\usepackage{placeins}
\usepackage{fvextra}

\newtcolorbox{questionbox}{%
  colback=gray!15, colframe=gray!15,
  boxrule=0pt, arc=2pt,
  left=4pt, right=4pt, top=2pt, bottom=2pt,
  before skip=4pt, after skip=4pt}
\newcommand{\Q}[2]{\begin{questionbox}\textbf{Q#1.}~#2\end{questionbox}}
\title{DataClawBench: An Agent Benchmark for Exploratory Real-World Financial Data Analysis}

\author{
  \textbf{Qiaohong Zhang}\textsuperscript{1}\thanks{\ Equal contribution.},
  \textbf{Weihao Ye}\textsuperscript{1}\footnotemark[1],
  \textbf{Jialong Chen}\textsuperscript{1},
  \textbf{Yi Luo}\textsuperscript{1},
  \\
  \textbf{BoYuan Li}\textsuperscript{1},
  \textbf{Bowen Deng}\textsuperscript{1},
  \textbf{Zibin Zheng}\textsuperscript{2},
  \textbf{Jianhao Lin}\textsuperscript{3},
  \\
  \textbf{Wei-Shi Zheng}\textsuperscript{1},
  \textbf{Chuan Chen}\textsuperscript{1}\thanks{\ Corresponding author.}
  \\
  \textsuperscript{1}\,School of Computer Science and Engineering, Sun Yat-sen University \\
  \textsuperscript{2}\,School of Software Engineering, Sun Yat-sen University\\
  \textsuperscript{3}\,Lingnan College, Sun Yat-sen University
}

\begin{document}
\maketitle
\etocdepthtag.toc{mainbody} 
\begin{abstract}
Autonomous data analysis agents are increasingly expected to conduct exploratory analysis with limited human guidance about data. However, existing benchmarks typically evaluate such agents in prior-guided settings, providing selected data sources, explicit data schemas, or cleaned data, thereby understating the exploratory burden. To evaluate this realistic exploratory data analysis task, we introduce DataClawBench, a benchmark built from financial think-tank consulting scenarios where agents must independently explore unfamiliar, noisy, cross-domain data and produce verifiable conclusions.
DataClawBench provides a unified real-world data environment with approximately 2.06 million records across enterprise, industry, and policy domains, with native data noise preserved. On top of this data environment, it defines 492 multi-step cross-domain tasks, each annotated with intermediate milestones that diagnose exploration and reasoning failures beyond outcome accuracy. A systematic evaluation of eight advanced LLMs under the OpenClaw agent reveals that exploratory data analysis breaks agent reliability: more exploration does not reliably translate into task-relevant progress or correct final answers\footnote{\url{https://huggingface.co/datasets/GTML-LAB/DataClaw}}.
\end{abstract}

\section{Introduction}
\label{sec:introduction}

\begin{table*}[t]
\centering
\small
\setlength{\tabcolsep}{6pt}
\renewcommand{\arraystretch}{1.15}

\scalebox{1.0}{
\begin{tabular}{lcccccc}
\toprule
\rowcolor{gray!20}
\textbf{Benchmark}
& \textbf{\begin{tabular}[c]{@{}c@{}}No Source\\Prior\end{tabular}}
& \textbf{\begin{tabular}[c]{@{}c@{}}No Schema\\Prior\end{tabular}}
& \textbf{\begin{tabular}[c]{@{}c@{}}No Noise\\Prior\end{tabular}}
& \textbf{\begin{tabular}[c]{@{}c@{}}Cross\\Domain\end{tabular}}
& \textbf{\begin{tabular}[c]{@{}c@{}}Process\\ Evaluation\end{tabular}}
& \textbf{\begin{tabular}[c]{@{}c@{}}Real-World\\Scenario\end{tabular}} \\
\midrule

Spider 2.0~\cite{spider2}
& $\times$ & $\times$ & $\times$ & $\times$ & $\times$ & $\checkmark$ \\
InfiAgent~\cite{infiagent}
& $\times$ & $\times$ & $\times$ & $\times$ & $\times$ & $\times$ \\
DACode~\cite{dacode}
& $\times$ & $\times$ & $\times$ & $\checkmark$ & $\times$ & $\checkmark$ \\
DABstep~\cite{dabstep}
& $\checkmark$ & $\checkmark$ & $\times$ & $\checkmark$ & $\times$ & $\checkmark$ \\
FDABench~\cite{fdabench}
& $\checkmark$ & $\checkmark$ & $\times$ & $\checkmark$ & $\times$ & $\times$ \\
FinanceBench~\cite{financebench}
& $\checkmark$ & $\checkmark$ & $\times$ & $\times$ & $\times$ & $\checkmark$ \\
\rowcolor{orange!10}\textbf{DataClawBench}
& $\checkmark$ & $\checkmark$ & $\checkmark$ & $\checkmark$ & $\checkmark$ & $\checkmark$ \\

\bottomrule
\end{tabular}
}
\caption{Comparison of representative benchmarks along key properties for exploratory data analysis.
The noise prior refers to either an explicit description of the noise or pre-cleaning of the data. $\checkmark$ denotes the presence of the corresponding key property, and $\times$ denotes its absence.}
\label{tab:benchmark-comparison}
\end{table*}

The rapid advancement of end-to-end agents driven by large language models (LLMs)~\cite{manus,openclaw} is reshaping data analysis. 
Traditionally, automated data analysis has often been formulated as static question answering, where a model directly produces an answer from specified text or tables. However, many real-world complex analytical tasks are exploratory before they are computational. Analysts often start with a concrete question, but the relevant evidence is not fully specified. They must inspect unfamiliar data environments, identify useful sources, align information across sources, handle data noise, and synthesize a conclusion~\cite{kandel2012enterprise,crisan2021baton}. As LLM agents become more autonomous, evaluating their ability to handle such exploratory tasks is increasingly important.

Existing benchmarks have made substantial progress in evaluating data analysis capabilities.  
However, these are not yet fully aligned with the requirements of exploratory real-world data analysis. Many benchmarks~\cite{infiagent, dacode,dabstep,fdabench,financebench} evaluate agents in prior-guided data settings, where relevant sources, schemas, or data noise conditions are often partly specified. 
This reduces the need for agents to discover evidence, align information across sources, and handle data noise. 
Yet in complex real-world analyses, such data priors are often unavailable or incomplete. 
As a result, they may overestimate how reliably agents can operate in real-world exploratory settings.

To fill this gap in evaluation, we need a realistic setting where task-relevant evidence is distributed across multiple data sources, with incomplete schema documentation and noise not fully known in advance. 
Financial think-tank consulting offers such a setting. 
In practice, analysts are asked to answer concrete consulting questions, such as enterprise diagnosis, industry comparison, and policy impact assessment, among others~\cite{oecd2015ddi,mckinsey2016analytics}. 
These tasks require analysts to connect evidence across operational records, industry indicators, and policy documents, turning imperfect data into verifiable conclusions~\cite{chu2016cleaning,hellerstein2018selfservice}. 
This makes it a natural testbed for evaluating exploratory data analysis under limited prior guidance.

Motivated by this setting, we propose DataClawBench, a benchmark for exploratory real-world financial data analysis. 
DataClawBench places agents in a unified underexplored data environment containing approximately 2.06 million real-world records across enterprise, industry, and policy domains, with native data noise preserved. 
It includes 492 expert-designed multi-step reasoning tasks derived from financial think-tank consulting scenarios. 
Each task is annotated with a unique final answer for objective evaluation, as well as intermediate milestones for diagnosing the agent's reasoning process. 
Table~\ref{tab:benchmark-comparison} compares DataClawBench with representative benchmarks along key properties for exploratory data analysis. 
Our contributions are summarized as follows.

\begin{itemize}
\item We introduce exploratory data analysis as a more challenging setting for evaluating autonomous data analysis agents. 
In this setting, agents must identify relevant evidence and handle data noise across multiple sources in an underexplored data environment.

\item We construct DataClawBench, a benchmark that instantiates this setting in financial think-tank consulting. 
DataClawBench restores the exploratory burden often abstracted away in prior-guided benchmarks.

\item We conduct a systematic evaluation of eight advanced LLMs under the OpenClaw. 
Seven models achieve below 50\% accuracy on DataClawBench. More exploration does not reliably translate into task-relevant progress.
\end{itemize}

\begin{figure*}[t]
\centering
\includegraphics[width=\textwidth]{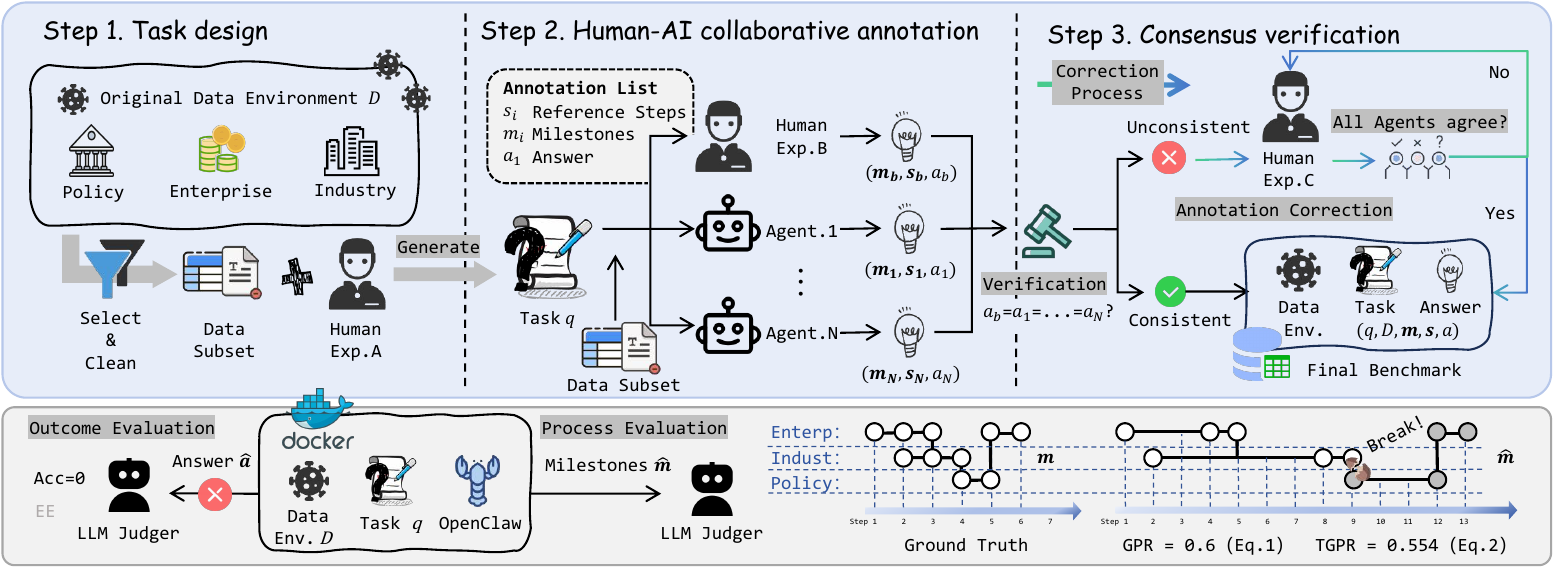}
\caption{Overall framework of DataClawBench. \textbf{Top}. Data annotation pipeline. \textbf{Bottom}. Evaluation pipeline. Each agent runs in an isolated Docker container, locates relevant information in an underexplored data environment, performs numerical computation and text comprehension, and produces a final answer, which is then assessed by both outcome evaluation and process evaluation.}
\label{fig:framework}
\end{figure*}

\section{Evaluation Protocol for Exploratory Data Analysis}
\label{sec:protocol}
Exploratory data analysis requires agents to first identify task-relevant evidence from heterogeneous data sources before deriving analytical conclusions. Unlike settings where sources, schemas, or cleaned tables are pre-specified, this process involves both evidence discovery and downstream reasoning. We therefore evaluate agents under a limited-prior setting, measuring not only answer correctness but also execution efficiency and exploratory progress.

\subsection{Problem setting.}
We formalize each task as a tuple
\(\mathcal{T} = (q, \mathcal{D}, \mathcal{P}, \mathbf{s}, \mathbf{m}, a)\),
where \(q\) is a natural-language analysis question and
\(\mathcal{D} = \{d_1, d_2, \dots, d_K\}\) denotes a data environment with \(K\) sources.
For each question, there exists a latent set of task-relevant evidence
\(\mathcal{R}_q \subseteq \mathcal{D}\), which is not directly exposed to the agent.
\(\mathcal{P}\) denotes task-level guidance priors available to the agent, such as source hints, schema descriptions, or noise descriptions.

DataClawBench instantiates a limited-prior evaluation setting. In contrast to prior-guided benchmarks that provide non-trivial priors \(\mathcal{P}\) to narrow the search space, DataClawBench withholds task-specific source hints, complete schema documentation, and explicit noise descriptions. Agents must therefore infer \(\mathcal{R}_q\) by exploring \(\mathcal{D}\).

The remaining components support verifiable evaluation. 
\(s=(s_1,\ldots,s_N)\) is an ordered gold reference trajectory of \(N\) core reasoning steps. 
\(m=\{(k_j,v_j)\}_{j=1}^{M}\) is a set of \(M\) milestone key-value pairs representing critical intermediate results. 
\(a\) is the gold answer. 
An agent \(\mathcal{A}\) receives \((q,\mathcal{D},\mathcal{P})\) and produces \(\mathcal{A}(q,\mathcal{D},\mathcal{P})=(\hat{s},\hat{a})\), where \(\hat{s}\) is the execution trajectory and \(\hat{a}\) is the predicted answer.

\subsection{Evaluation metrics.}
We evaluate agents along three complementary dimensions: effectiveness, efficiency, and exploratory progress.

\paragraph{Effectiveness.}
Effectiveness measures whether the agent reaches the correct analytical conclusion. We use final answer correctness (\textbf{Acc}), evaluated by an LLM judge that compares the predicted answer \(\hat{a}\) with the gold answer \(a\). For tasks containing \(L\) sub-questions, the judge scores each sub-question independently and reports the normalized accuracy in \([0,1]\). The judging prompt is provided in Appendix~\ref{appendix:prompt}.

\paragraph{Efficiency.}
Efficiency measures whether the agent reaches a correct conclusion with limited redundant exploration. For correctly answered tasks, we define execution efficiency (\textbf{EE}) as $\eta = \frac{N}{T}$,
where \(N=|\mathbf{s}|\) is the gold trajectory length and \(T=|\hat{\mathbf{s}}|\) is the number of agent execution steps. An agent step is defined as one round of model inference together with the corresponding tool-result return, and may contain multiple tool calls. Higher EE indicates less redundant exploration.

\paragraph{Exploratory progress.}
Exploratory progress measures how much task-relevant progress an agent makes when the final answer is incorrect. We evaluate it using two milestone-based metrics. \textbf{Goal Progress Rate (GPR)} measures the fraction of gold milestones achieved during execution:
\begin{equation}
\mathrm{GPR} = \frac{1}{M} \sum_{j=1}^{M} \mathbb{I}(m_j),
\label{eq:gpr}
\end{equation}
where \(M\) is the number of gold milestones, and \(\mathbb{I}(m_j)=1\) if milestone \(m_j\) is correctly achieved in the predicted trajectory \(\hat{\mathbf{s}}\). Milestone achievement is judged by an LLM based on either direct evidence in the trajectory or indirect evidence from correctly completed downstream milestones.

\textbf{Temporal Progress Efficiency (TPE)} further captures whether achieved milestones are reached in a timely manner:
\begin{equation}
\mathrm{TPE} = 
\frac{\sum_{j=1}^{M} \mathbb{I}(m_j) \cdot \gamma^{\max(t_j - N,\; 0)}}
{\sum_{j=1}^{M} \mathbb{I}(m_j)}
\;\in\; [0, 1],
\label{eq:tpe}
\end{equation}
where \(t_j\) is the first step at which milestone \(m_j\) is achieved, and \(\gamma \in (0,1]\) is the temporal decay factor. Milestones achieved within the first \(N\) steps receive full credit, while later milestones are exponentially discounted. Thus, \(\mathrm{TPE}=1\) indicates that all achieved milestones are reached on time, whereas lower values indicate delayed progress.

\begin{table*}[t]
\centering
\small
\resizebox{\textwidth}{!}{
\begin{tabular}{
>{\raggedright\arraybackslash}m{2.8cm}
>{\raggedright\arraybackslash}m{7.8cm}
>{\raggedright\arraybackslash}m{1.8cm}
>{\raggedright\arraybackslash}m{0.9cm}
>{\raggedright\arraybackslash}m{1.0cm}
}
\toprule
\rowcolor{gray!18}
\textbf{Category} & \textbf{Example Question} & \textbf{E / M / H} & \textbf{Total} & \textbf{Steps} \\
\midrule
\rowcolor{gray!8}
\makecell[l]{Enterprise--Industry \\ Analysis}
& \textit{In 2022, is the operating revenue of Company X higher than the total operating revenue of its industry in its province?}
& 115 / 111 / -- & 226 & 2--5 \\
\midrule

\makecell[l]{Enterprise--Industry-- \\ Policy Analysis}
& \textit{Compare the number of central-level policies for Company A's industry with local policies for Company B's industry.}
& 10 / 66 / -- & 76 & 3--5 \\
\midrule

\rowcolor{gray!8}
Comprehensive Decision
& \textit{In 2022, an automotive manufacturer scored each province's industrial supporting capacity before selecting a plant site. The scoring rules are as follows...what is the composite index value of the province with the highest industrial supporting composite index?}
& 6 / 45 / 19 & 70 & 2--8 \\
\midrule

International Comparison
& \textit{What is the ratio of Futu Holdings' net profit per employee to the median of China's capital market services industry?}
& -- / 25 / 14 & 39 & 4--7 \\
\midrule

\rowcolor{gray!8}
Hypothesis Verification
& \textit{Verify whether enterprise total assets correlate with invention patent count in Consumer Electronics, restricted to provinces with above-average R\&D density.}
& -- / 14 / 15 & 29 & 5--9 \\
\midrule

Industry Planning
& \textit{For Guangdong's consumer electronics industry, compare high-end and export-oriented strategies using inter-provincial ranking scores.}
& -- / 14 / 14 & 28 & 5--9 \\
\midrule

\rowcolor{gray!8}
Risk Assessment
& \textit{If all national R\&D tax incentives are cancelled, which three manufacturing industries suffer the largest average net profit margin decline?}
& -- / 11 / 13 & 24 & 5--9 \\
\midrule

\textbf{Total}
&
& \textbf{131 / 286 / 75}
& \textbf{492}
& \textbf{2--9} \\
\bottomrule
\end{tabular}}
\caption{Task distribution by category and difficulty level. Steps denotes the range of gold reference step counts. Due to space constraints, the example questions shown are abbreviated or partial versions of the actual questions.}
\label{tab:task-categories}
\end{table*}

\begin{table}[t]
\centering
\resizebox{\columnwidth}{!}{
\begin{tabular}{lllr}
\toprule
\rowcolor{gray!18}
\textbf{Domain} & \textbf{Source} & \textbf{Type} & \textbf{Scale} \\
\midrule
\multirow{3}{*}{Enterprise} & Company profiles & Struct. & \multirow{3}{*}{1,820K} \\
 & Company operations & Struct. & \\
 & Core competitiveness & Unstruct. & \\
\midrule
\multirow{2}{*}{Industry} & National statistics & Struct. & \multirow{2}{*}{234K} \\
 & Regional statistics & Struct. & \\
\midrule
\multirow{2}{*}{Policy} & Policy documents & Mixed & \multirow{2}{*}{9K} \\
 & Release statistics & Struct. & \\
\midrule
\multicolumn{2}{l}{Internal knowledge base} & Unstruct. & 25 \\
\bottomrule
\end{tabular}}
\caption{DataClawBench data environment. Seven core data sources span three domains, expanding into 18 tables. Scale is measured in number of records.}
\label{tab:data-env}
\end{table}

\section{DataClawBench}
\label{sec:DataClawBench}

\subsection{Underexplored Real-World Data Environment}
\label{subsec:data-env}
\paragraph{Data source composition.}
We construct an underexplored real-world data environment $\mathcal{D}$ with approximately 2.06 million records from seven core data sources spanning enterprise, industry, and policy domains. Table~\ref{tab:data-env} summarizes the source composition and data types, while Appendix Table~\ref{tab:data-full} provides more details.
The enterprise domain includes publicly listed technology-oriented companies and expert-selected key unlisted technology firms. The collected data cover static company profiles, annual financial and operating indicators, and textual descriptions of core competitiveness. The industry domain captures national- and provincial-level statistics on industrial economic performance, scale, and patent activity. The policy domain consists of full policy texts and publication metadata related to science and technology innovation at both national and local levels.

\paragraph{Data characteristics.} These sources form a cross-domain data network connected by implicit business attributes, such as registration location and industry classification. This structure enables the benchmark to link micro-level firm operations, meso-level industry patterns, and macro-level policy environments. Rather than curating the data into a simplified benchmark setting, DataClawBench deliberately retains real-world data imperfections, such as missing values, inconsistent units, ambiguous entity names, and multilingual expressions. Moreover, data schemas are not explicitly provided across sources. Some can be partially inferred from table headers, whereas others are difficult to recover due to irregular table structures.

\paragraph{Data provenance and anonymization.}
The core data of DataClawBench originates from the publishing team's long-term think-tank research and consulting practice, rather than synthetic samples or textbook examples. To preserve reasoning value while protecting commercial privacy, DataClawBench employs a three-stage anonymization pipeline. Full details are provided in Appendix~\ref{appendix:anonymization}

\subsection{Task Design and Construction}
\label{subsec:construction}

\paragraph{Task composition.}
DataClawBench consists of 492 data analysis tasks designed by domain experts. As shown in Table~\ref{tab:task-categories}, these tasks are organized into seven categories according to thematic scope. Most tasks involve joint analysis of micro-level enterprise performance, meso-level industry trends, and macro-level policy contexts, while international comparison tasks additionally require web searches for external information.
Moreover, the tasks are designed to require multi-step reasoning, where intermediate results are often used as inputs to subsequent steps. Each task has a unique answer, supporting deterministic evaluation. Because of the required domain knowledge and annotation complexity, a human expert may spend more than one hour annotating a single task.

\paragraph{Difficulty annotation.}
We adopt different difficulty annotation strategies for numerical and text-intensive tasks. For the first four categories, which mainly involve numerical computation, difficulty is based on the number of annotated reasoning steps: 2--3 steps for easy, 4--5 for medium, and 6 or more for hard. For the last three categories, which involve long policy, difficulty is manually assigned by domain experts because step count alone cannot capture the reading and reasoning burden.

\paragraph{Human-in-the-loop annotation pipeline.} DataClawBench adopts a three-stage human-in-the-loop annotation pipeline, consisting of expert task design, human-AI collaborative annotation, and consensus verification. This design combines domain expertise with model-assisted efficiency. The complete process is illustrated in Figure~\ref{fig:framework}.

\ding{182}~\textbf{Task design.} Domain Expert~A first constructs a cleaned and idealized version of the data environment, and then designs analysis tasks based on it.
\ding{183}~\textbf{Human-AI collaborative annotation.} In the cleaned environment, Domain Expert~B and multiple data analysis agents independently solve the same task. Each participant produces a candidate answer $a$, a reasoning trace $\mathbf{s}$, and milestone annotations $\mathbf{m}$. These parallel solutions provide references for subsequent verification.
\ding{184}~\textbf{Consensus verification.} Domain Expert~C compares the human and agent outputs. If all participants reach the same final answer, the human annotation is directly accepted. Otherwise, Expert~C examines the conflicting reasoning paths, corrects the annotation when necessary, and iterates the process until a consensus annotation is obtained.

This consensus-based verification reduces the risk of relying on a single annotator or model. To keep annotation feasible, the process is conducted in the cleaned data environment, so the final annotations focus on core solution steps rather than exploratory trial-and-error.

\section{Experiments}
\label{sec:experiments}

We evaluate eight agents driven by LLMs spanning different model families and scales. 
All experiments use the OpenClaw agent framework.
Each task is executed in an isolated Docker container with the full data environment $\mathcal{D}$ mounted as a read-only workspace.
Agents may use any tools available in OpenClaw, including Python execution, file reading, web fetch, and web search, to explore the data and derive answers.
A per-task timeout of 1200 seconds is enforced to bound computation. All evaluation judging is performed by GLM-5. Further experimental details are provided in Appendix~\ref{appendix:exp-details}.

\begin{table*}[t]
\centering
\small
\renewcommand{\arraystretch}{1.12}
\setlength{\tabcolsep}{6pt}
\begin{tabular}{@{}l cc cc >{\columncolor{orange!10}}c cc >{\columncolor{orange!10}}c cc@{}}
\toprule
\rowcolor{gray!18}
 & \multicolumn{2}{c}{\textbf{Easy}}
 & \multicolumn{2}{c}{\textbf{Medium}}
 & \multicolumn{1}{>{\columncolor{gray!18}}c}{\textbf{$\Delta$ Acc.}}
 & \multicolumn{2}{c}{\textbf{Hard}}
 & \multicolumn{1}{>{\columncolor{gray!18}}c}{\textbf{$\Delta$ Acc.}}
 & \multicolumn{2}{c}{\textbf{Overall}} \\

\rowcolor{gray!18}
\multirow{-2}{*}{\textbf{Model}}
 & Acc.\,($\uparrow$) & EE\,($\uparrow$)
 & Acc.\,($\uparrow$) & EE\,($\uparrow$)
 & E$\to$M\
 & Acc.\,($\uparrow$) & EE\,($\uparrow$)
 & M$\to$H\
 & Acc.\,($\uparrow$) & EE\,($\uparrow$) \\

\midrule
Claude Opus 4.6
 & \textbf{76.8} & 0.41
 & \textbf{62.2} & 0.43
 & 14.6
 & \textbf{44.8} & 0.44
 & 17.4
 & \textbf{63.4} & 0.42 \\

Gemini 3.1 Pro$^\dagger$
 & 67.6 & 0.42
 & 44.6 & 0.26
 & 23.0
 & 12.3 & 0.30
 & 32.3
 & 45.8 & 0.32 \\

Minimax M2.7
 & 62.6 & 0.31
 & 37.2 & 0.21
 & 25.4
 & 19.3 & 0.26
 & 17.9
 & 41.3 & 0.26 \\

Qwen3.5-Plus
 & 45.0 & 0.31
 & 40.0 & 0.20
 & 5.0
 & 16.2 & 0.29
 & 23.8
 & 37.7 & 0.24 \\

DeepSeek-V3.2
 & 45.0 & 0.15
 & 35.8 & 0.15
 & 9.2
 & 8.9 & 0.20
 & 26.9
 & 34.1 & 0.16 \\

GLM-5
 & 45.8 & 0.32
 & 33.0 & 0.32
 & 12.8
 & 12.6 & 0.30
 & 20.4
 & 33.3 & 0.32 \\

Kimi-K2.5
 & 41.2 & 0.37
 & 22.6 & 0.31
 & 18.6
 & 14.0 & 0.26
 & 8.6
 & 26.3 & 0.33 \\

GPT-5.4
 & 23.1 & \textbf{0.45}
 & 26.5 & \textbf{0.46}
 & $-$3.4
 & 12.0 & \textbf{0.59}
 & 14.5
 & 23.4 & \textbf{0.46} \\

\midrule
\rowcolor{gray!12}
\textit{Mean}
 & 50.9 & 0.34
 & 37.7 & 0.29
 & 13.2
 & 17.5 & 0.33
 & 20.2
 & 38.2 & 0.31 \\
\bottomrule
\end{tabular}

\caption{Outcome scores. Arrows indicate the preferred direction of each metric. The two $\Delta$ columns report adjacent difficulty-stage accuracy drops. $^\dagger$Gemini 3.1 Pro denotes Gemini 3.1 Pro Preview.}
\label{tab:outcome}
\end{table*}

\begin{table*}[t]
\centering
\small
\renewcommand{\arraystretch}{1.12}
\setlength{\tabcolsep}{5.2pt}
\begin{tabular}{@{}l cc >{\columncolor{orange!10}}c cc >{\columncolor{orange!10}}c cc >{\columncolor{orange!10}}c cc@{}}
\toprule
\rowcolor{gray!18}
 & \multicolumn{2}{c}{\textbf{Easy}}
 & \multicolumn{1}{>{\columncolor{gray!18}}c}{\textbf{$\Delta$}}
 & \multicolumn{2}{c}{\textbf{Medium}}
 & \multicolumn{1}{>{\columncolor{gray!18}}c}{\textbf{$\Delta$}}
 & \multicolumn{2}{c}{\textbf{Hard}}
 & \multicolumn{1}{>{\columncolor{gray!18}}c}{\textbf{$\Delta$}}
 & \multicolumn{2}{c}{\textbf{Overall}} \\

\rowcolor{gray!18}
\multirow{-2}{*}{\textbf{Model}}
 & GPR\,($\uparrow$) & TPE\,($\uparrow$)
 & \textbf{Rank}
 & GPR\,($\uparrow$) & TPE\,($\uparrow$)
 & \textbf{Rank}
 & GPR\,($\uparrow$) & TPE\,($\uparrow$)
 & \textbf{Rank}
 & GPR & TPE \\
\midrule

Claude Opus 4.6
 & \textbf{42.9} & 0.56
 & 0
 & \textbf{45.6} & 0.63
 & 0
 & \textbf{45.4} & 0.50
 & 0
 & \textbf{45.1} & 0.59 \\

Gemini 3.1 Pro$^\dagger$
 & 17.2 & 0.54
 & 0
 & 40.1 & 0.43
 & 0
 & 28.7 & 0.26
 & \cellcolor{green!20}$4$
 & 33.6 & 0.41 \\

Minimax M2.7
 & 16.3 & 0.40
 & $-1$
 & 39.9 & 0.37
 & $1$
 & 26.6 & 0.36
 & $-1$
 & 33.2 & 0.37 \\

Qwen3.5-Plus
 & 11.7 & 0.29
 & 0
 & 24.9 & 0.37
 & $-2$
 & 24.0 & 0.42
 & $-1$
 & 21.7 & 0.37 \\

DeepSeek-V3.2
 & 16.4 & 0.36
 & \cellcolor{green!20}$3$
 & 28.7 & 0.33
 & $1$
 & 22.4 & 0.34
 & \cellcolor{green!20}$3$
 & 24.7 & 0.34 \\

GLM-5
 & 7.3 & 0.38
 & \cellcolor{red!15}$-3$
 & 23.8 & 0.53
 & $-1$
 & 20.0 & 0.54
 & $-1$
 & 19.5 & 0.52 \\

Kimi-K2.5
 & 9.5 & 0.59
 & $1$
 & 23.7 & 0.56
 & 0
 & 15.0 & 0.61
 & \cellcolor{red!15}$-4$
 & 19.2 & 0.57 \\

GPT-5.4
 & 6.0 & \textbf{0.68}
 & 0
 & 24.2 & \textbf{0.76}
 & $1$
 & 18.8 & \textbf{0.73}
 & 0
 & 18.5 & \textbf{0.75} \\

\midrule
\rowcolor{gray!12}
\textit{Mean}
 & 15.9 & 0.48
 & ---
 & 31.4 & 0.50
 & ---
 & 25.1 & 0.47
 & ---
 & 26.9 & 0.49 \\

\bottomrule
\end{tabular}

\caption{Process scores on incorrectly answered tasks with $\gamma=0.9$. $\Delta$Rank measures the difficulty-specific rank divergence between final accuracy and goal progress, computed as the Acc rank minus the GPR rank.}
\label{tab:process}
\end{table*}

\subsection{Main Results}

\Q{1}{Can current agents remain reliable when they must find the relevant data by themselves?}
\textbf{Not reliably.} Table~\ref{tab:outcome} summarises outcome scores across all 492 tasks. Even the strongest agent, Claude Opus 4.6, reaches only 63.4\% overall accuracy. 
The other seven models all fall below 50\%, and five of the eight fall below 40\%. 
No model approaches saturation, indicating that exploratory analysis over real-world data environments remains far from solved. 
Performance also deteriorates sharply as task complexity increases. Averaged across the eight models, accuracy drops by 13.2 percentage points from Easy to Medium, but by 20.2 percentage points from Medium to Hard, a decline more than $1.5\times$ larger. Claude Opus 4.6 drops from 76.8\% on Easy tasks to 44.8\% on Hard tasks, while other models show even steeper declines, with the largest Acc drop reaching 55.3 percentage points. Per-category results for all eight models are provided in Appendix~\ref{appendix:cross-task}.

\begin{figure*}[t]
\centering
\begin{subfigure}[t]{0.32\textwidth}
\centering
\includegraphics[width=\linewidth]{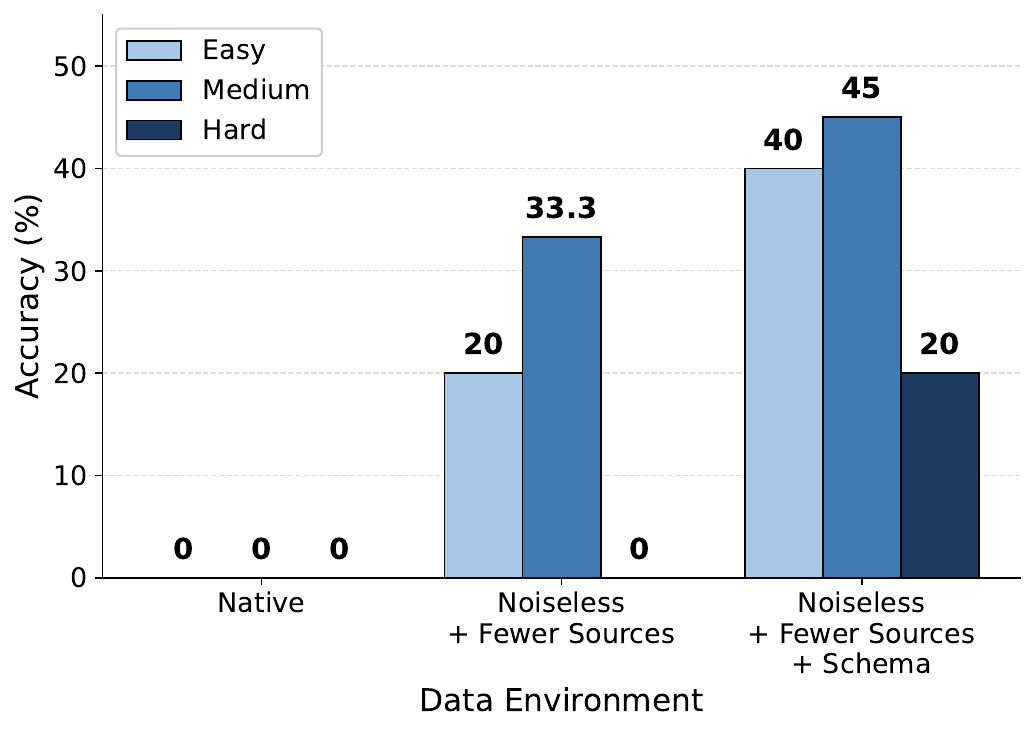}
\caption{Acc under different environments.}
\label{fig:main-cleaning}
\end{subfigure}\hfill
\begin{subfigure}[t]{0.32\textwidth}
\centering
\includegraphics[width=\linewidth]{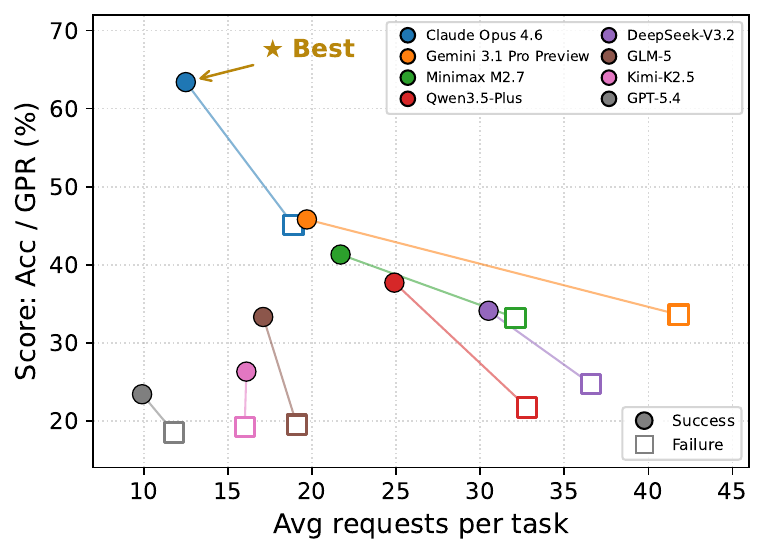}
\caption{Avg requests on failed tasks vs.\ GPR.}
\label{fig:main-reqs-gpr}
\end{subfigure}\hfill
\begin{subfigure}[t]{0.32\textwidth}
\centering
\includegraphics[width=\linewidth]{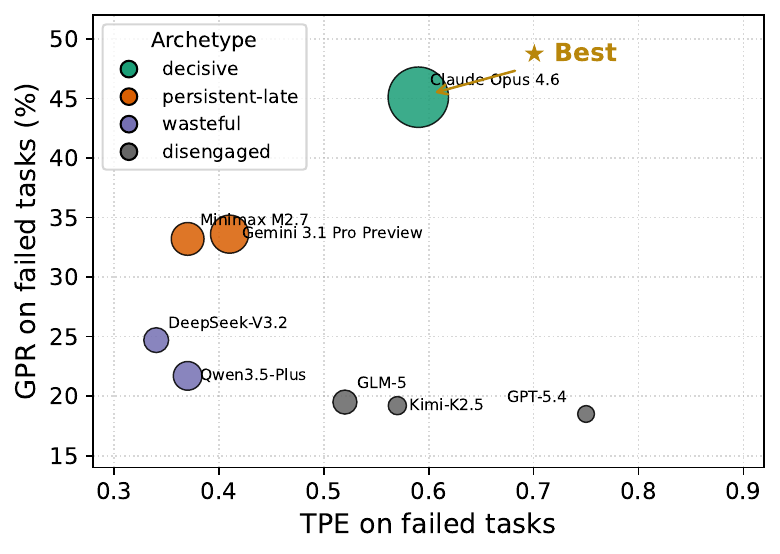}
\caption{TPE vs.\ GPR.}
\label{fig:main-bubble}
\end{subfigure}
\caption{Three diagnostic views of agent behaviour on DataClawBench. 
(c) The eight models partition into four exploration archetypes, and outcome accuracy (bubble size) co-varies with this partition. Color $=$ archetype.}
\label{fig:main-3panel}
\end{figure*}

\Q{2}{Is longer exploration actually converted into useful progress?}

\textbf{Only partially.} As shown in Figure~\ref{fig:main-reqs-gpr}, failed tasks with longer exploration generally achieve higher GPR, indicating that additional requests can recover some intermediate progress. The early-stopping models, including GPT-5.4, Kimi-K2.5, and GLM-5, use only about 12--19 requests on failed tasks and reach roughly 18--20\% GPR. By contrast, Gemini 3.1 Pro, Minimax M2.7, DeepSeek-V3.2, and Qwen3.5-Plus explore longer, using about 32--42 requests, and achieve higher GPR of around 22--34\%.

However, more requests do not translate into progress uniformly. Claude achieves the highest GPR with relatively few failed-task requests, whereas Gemini and Minimax spend substantially more requests but obtain lower progress. This suggests that exploration length alone is a weak proxy for useful progress. What matters is whether exploration remains goal-directed rather than longer.

\Q{3}{What failure modes are hidden behind the same wrong answer?}
\textbf{The same wrong answer can hide very different process failure modes.} Final accuracy only tells us whether an answer is correct. It does not show how the agent failed. The gap between Acc rank and GPR rank shows this clearly. Table~\ref{tab:process} reports that on Hard tasks, Gemini 3.1 Pro ranks sixth by final accuracy but second by GPR. This means that many of its failed runs still reach useful intermediate milestones. Kimi-K2.5 shows the opposite pattern. It ranks fourth by accuracy but last by GPR, which means that its failed runs often reach fewer milestones. From the outcome view, these two agents may look comparable. From the process view, they are fundamentally different agents.

\Q{4}{When agents fail, what exploration profiles do they exhibit?}
\label{subsubsec:efficiency}

\textbf{Combining Acc, GPR, and TPE reveals four exploration profiles.} 
In Figure~\ref{fig:main-bubble}, the decisive solver (Claude Opus 4.6) achieves high GPR without sacrificing TPE, reaching a substantial fraction of milestones early. It also has high Acc. The persistent but late archetype (Gemini 3.1 Pro Preview, Minimax M2.7) holds high GPR but low TPE, accumulating milestones eventually rather than early. The wasteful trial-and-error archetype (DeepSeek-V3.2, Qwen3.5-Plus) has moderate GPR and low TPE, spending many steps without proportionate progress. The disengaged archetype (GPT-5.4, Kimi-K2.5, GLM-5) has the lowest GPR but high TPE, they terminate early, so the few milestones they reach come early. 

\subsection{Failure Attribution}

\Q{5}{What makes exploratory data analysis difficult: noisy data or missing guidance?}
\label{subsubsec:claim2}

\textbf{Both matter.} We re-evaluate Qwen3.5-Plus on 30 originally failed tasks under three data environments with decreasing environmental uncertainty, as shown in Figure~\ref{fig:main-cleaning}. The first is the original real-world environment. The second removes data noise and irrelevant sources. The third further provides complete schema guidance on top of the second environment. 
Reducing environmental uncertainty improves accuracy, but the gains are uneven across difficulty levels. In the original environment, model fails on all sampled tasks. After removing noise and irrelevant sources, accuracy recovers to 20\% on Easy and 33.3\% on Medium, but remains 0\% on Hard tasks. When complete schema guidance is added, accuracy further rises to 40\% on Easy, 45\% on Medium, and 20\% on Hard tasks.

This shows that exploratory data analysis is not hard only because the data is noisy. Removing noise and irrelevant sources helps agents enter a more useful evidence space, especially on Easy and Medium tasks. However, Hard remain difficult until stronger schema guidance is provided, and even then recover only partially. This suggests that agents struggle with multiple forms of uncertainty at once: noisy evidence, irrelevant sources, and incomplete schema understanding.
Other models show consistent trends, detailed in Appendix~\ref{appendix:env-impact}.

\begin{figure}[t]
\centering
\includegraphics[width=\linewidth]{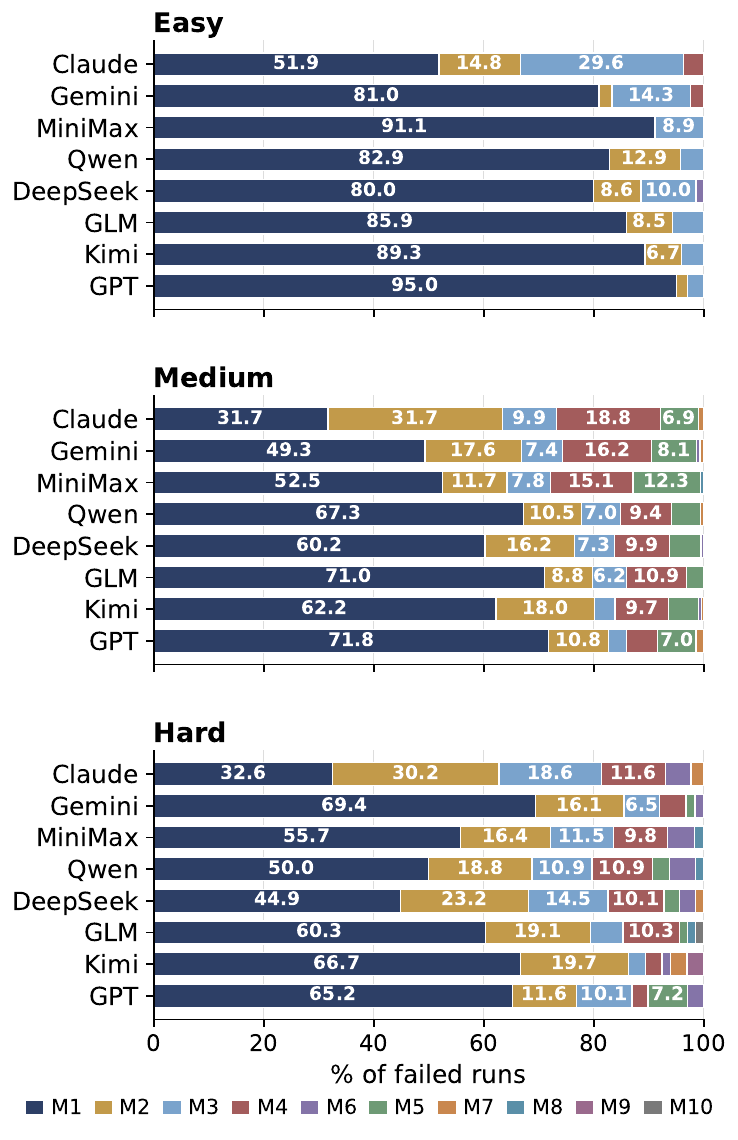}
\caption{Position $m_k$ of the first un-achieved milestone, shown separately for Easy, Medium, and Hard tasks.}
\label{fig:fail_step_distribution}
\end{figure}

\Q{6}{Where do agents first lose the analytical thread?}
\textbf{Agents usually lose the analytical thread early, but stronger agents fail later than weaker ones.} We localize the origin of each failed run by recording the first unachieved gold milestone $m_k$. 
Figure~\ref{fig:fail_step_distribution} reports this distribution for all eight agents, grouped by difficulty. 
Failures are concentrated near the beginning of the analytical chain, but the depth of failure differs across models. 
Claude Opus 4.6, the strongest agent, is less likely to collapse at the very first milestone. 
For most agents, the first milestone $M_1$ accounts for a large share of failures. 
This pattern suggests that many agents derail before establishing the initial evidence base, framing the problem, or setting up intermediate variables. Stronger agents can often move beyond the initial evidence-acquisition stage and fail later.

\begin{table*}[t]
\centering
\small
\setlength{\tabcolsep}{11pt}
\begin{tabular}{l ccccc}
\toprule
\textbf{First-failing operation} 
& \makecell[c]{\textbf{Wrong-Answer}\\\textbf{Stop}} 
& \makecell[c]{\textbf{Voluntary}\\\textbf{Give-up}} 
& \makecell[c]{\textbf{Silent}\\\textbf{Stop}} 
& \makecell[c]{\textbf{Timeout}\\\textbf{Kill}} 
& \textbf{Other} \\
\midrule
Entity Attribute Lookup        & 37.2                   & \cellcolor{green!20}45.7 & 10.3 &  5.3 & 1.4 \\
Aggregate Count or Sum         & \cellcolor{red!15}69.0 & 10.6                     & 10.6 &  8.8 & 1.0 \\
Statistical Summary            & 56.8                   & 22.2                     &  9.0 & 11.5 & 0.4 \\
Policy Lookup and Count        & \cellcolor{red!15}66.0 & 16.9                     &  5.0 & 10.6 & 1.6 \\
Comparison or Boolean Judgment & \cellcolor{red!15}63.1 & 15.5                     &  5.3 & 13.1 & 2.9 \\
Ranking and Selection          & 56.9                   & 25.0                     &  7.5 &  6.9 & 3.8 \\
\bottomrule
\end{tabular}
\caption{Termination behavior by first-failing operation. Category definitions are provided in Appendix~\ref{appendix:operation-taxonomy}.}
\label{tab:op-by-term}
\end{table*}

\Q{7}{What makes agents lose the thread?}
\textbf{Whether an agent gives up or commits to a wrong answer tracks the operation type at the breakpoint.} Table~\ref{tab:op-by-term} cross-tabulates failed runs by the operation type of the first un-achieved milestone and the final termination mode. 
A clear divide emerges between concrete lookup failures and complex evidence failures. 
For Entity Attribute Lookup, Voluntary Give-up is the largest termination mode at 45.7\%, suggesting that agents can often recognize missing concrete evidence, such as a row, field, or entity. 
By contrast, Wrong-Answer Stop dominates operations that can still yield plausible outputs despite incomplete or misused evidence: Aggregate Count or Sum, Policy Lookup and Count, and Comparison or Boolean Judgment reach 69.0\%, 66.0\%, and 63.1\%, respectively, while Ranking and Selection and Statistical Summary also exceed 56\%. 
These operations rarely provide a clean ``no result'' signal, so agents tend to commit to a number, ranking, comparison, or summary rather than admit the gap. 

Easy failures are dominated by concrete lookup and comparison, where give-up remains available.
Hard failures involve more policy, aggregation, statistical, and ranking operations, where agents are more likely to overcommit. 
Thus, agents lose the analytical thread not only because they perform the wrong operation, but also because they choose the wrong stopping action after the operation fails.

\section{Related Work}
\label{sec:related}

\paragraph{Data Analysis under Prior-Guided Data Settings.}  Early research focused on static question answering progressed from single-table QA~\cite{wikitablequestions,spider} to cross-database text-to-SQL~\cite{spider2}, and from multi-hop textual reasoning~\cite{hotpotqa,feb4rag} to hybrid reasoning over tables and text~\cite{finqa,convfinqa}. Recent work has begun to explore agent-oriented evaluation of multi-step data analysis. DA-Code~\cite{dacode} and InfiAgent-DABench~\cite{infiagent} require agents to write code under explicitly stated data sources. FinanceBench~\cite{financebench} targets financial analysis over standardized SEC filings.
Despite these advances in task complexity and data scale, many of them still provide partial guidance about data sources, schemas, or data noise conditions. Most of these benchmarks therefore primarily evaluate reasoning over prior-guided data rather than analytical capability in underexplored environments.

\paragraph{Outcome-Oriented Evaluation Paradigms.} Most agent benchmarks~\cite{agentbench, webarena, swe-ci} adopt an outcome-only evaluation approach, treating the reasoning process as a black box. Recent data analysis works~\cite{dabstep, deepresearchqa} explicitly identify process-level evaluation as an important future direction. A few recent works also have begun to explore process-level evaluation. AgentBoard~\cite{agentboard} introduces a progress rate metric for incremental task completion, and Galileo~\cite{galileo} proposes an action advancement measure to assess whether each step progresses toward the goal.  Unlike the operation-step-level progress metrics above, DataClawBench evaluates goal achievement at the level of critical intermediate results, decoupling evaluation from specific execution paths to accommodate diverse analytical strategies while localizing semantic break points in the reasoning chain.

\section{Conclusion}
\label{sec:conclusion}

We present DataClawBench, a benchmark that evaluates LLM-driven agents on financial data analysis tasks in underexplored real-world data environments. It withholds source and schema priors, preserves native data noise, and annotates tasks with process-level milestones. DataClawBench reveals not only whether an agent succeeds but how and where it fails. Experiments show that advanced LLMs struggle on DataClawBench and different models adopt markedly different exploration styles. Overall, DataClawBench provides a diagnostic testbed for probing the capability boundaries of autonomous financial data-analysis agents.

\section*{Limitations and Future work}
DataClawBench currently focuses on structured and unstructured textual modalities. Extending to richer modalities would better reflect real-world analytical workflows.
Additionally, the milestone annotations are produced through human-in-the-loop pipelines, meaning that task decomposition and labeling granularity are inevitably subject to annotator judgment. The resulting milestones should therefore be treated as one plausible reference rather than a unique ground truth. Finally, A limitation of our current process evaluation is that GPR and TPE are computed only on tasks with incorrect final answers. As a result, these metrics do not fully capture cases where an agent arrives at the correct answer through incomplete, accidental, or flawed reasoning. Evaluating such cases would require more fine-grained and rigorous process-level assessment, which we leave for future work.

\bibliography{dataclaw/ref}

@String{Computing = "Computing" }

@String{Computer = "{IEEE} Computer" }

@article{manus,
  title={From mind to machine: The rise of manus ai as a fully autonomous digital agent},
  author={Shen, Minjie and Li, Yanshu and Chen, Lulu and Fan, Zhichao and Li, Yanhang and Yang, Qikai},
  journal={arXiv preprint arXiv:2505.02024},
  year={2025}
}

@misc{openclaw,
  author    = {Peter Steinberger},
  title     = {{OpenClaw}: An Open-Source Autonomous {AI} Agent},
  year      = {2025},
  url       = {https://github.com/openclaw/openclaw}
}

@inproceedings{wikitablequestions,
  title={Compositional semantic parsing on semi-structured tables},
  author={Pasupat, Panupong and Liang, Percy},
  booktitle={Proceedings of the 53rd Annual Meeting of the Association for Computational Linguistics and the 7th International Joint Conference on Natural Language Processing (Volume 1: Long Papers)},
  pages={1470--1480},
  year={2015}
}

@inproceedings{spider,
 title = "{S}pider: A Large-Scale Human-Labeled Dataset for Complex and Cross-Domain Semantic Parsing and Text-to-{SQL} Task",
    author = "Yu, Tao  and
      Zhang, Rui  and
      Yang, Kai  and
      Yasunaga, Michihiro  and
      Wang, Dongxu  and
      Li, Zifan  and
      Ma, James  and
      Li, Irene  and
      Yao, Qingning  and
      Roman, Shanelle  and
      Zhang, Zilin  and
      Radev, Dragomir",
    editor = "Riloff, Ellen  and
      Chiang, David  and
      Hockenmaier, Julia  and
      Tsujii, Jun{'}ichi",
    booktitle = "Proceedings of the 2018 Conference on Empirical Methods in Natural Language Processing",
    month = oct # "-" # nov,
    year = "2018",
    address = "Brussels, Belgium",
    publisher = "Association for Computational Linguistics",
    url = "https://aclanthology.org/D18-1425/",
    doi = "10.18653/v1/D18-1425",
    pages = "3911--3921",
}

@inproceedings{hotpotqa,
  title = "{H}otpot{QA}: A Dataset for Diverse, Explainable Multi-hop Question Answering",
    author = "Yang, Zhilin  and
      Qi, Peng  and
      Zhang, Saizheng  and
      Bengio, Yoshua  and
      Cohen, William  and
      Salakhutdinov, Ruslan  and
      Manning, Christopher D.",
    editor = "Riloff, Ellen  and
      Chiang, David  and
      Hockenmaier, Julia  and
      Tsujii, Jun{'}ichi",
    booktitle = "Proceedings of the 2018 Conference on Empirical Methods in Natural Language Processing",
    month = oct # "-" # nov,
    year = "2018",
    address = "Brussels, Belgium",
    publisher = "Association for Computational Linguistics",
    url = "https://aclanthology.org/D18-1259/",
    doi = "10.18653/v1/D18-1259",
    pages = "2369--2380"
}

@inproceedings{finqa,
  title={Finqa: A dataset of numerical reasoning over financial data},
  author={Chen, Zhiyu and Chen, Wenhu and Smiley, Charese and Shah, Sameena and Borova, Iana and Langdon, Dylan and Moussa, Reema and Beane, Matt and Huang, Ting-Hao and Routledge, Bryan R and Wang, William Yang},
  booktitle={Proceedings of the 2021 Conference on Empirical Methods in Natural Language Processing},
  pages={3697--3711},
  year={2021}
}

@inproceedings{convfinqa,
  title={Convfinqa: Exploring the chain of numerical reasoning in conversational finance question answering},
  author={Chen, Zhiyu and Li, Shiyang and Smiley, Charese and Ma, Zhiqiang and Shah, Sameena and Wang, William Yang},
  booktitle={Proceedings of the 2022 Conference on Empirical Methods in Natural Language Processing},
  pages={6279--6292},
  year={2022}
}

@article{financebench,
  title={Finance agent benchmark: Benchmarking llms on real-world financial research tasks},
  author={Bigeard, Antoine and Nashold, Langston and Krishnan, Rayan and Wu, Shirley},
  journal={arXiv preprint arXiv:2508.00828},
  year={2025}
}

@inproceedings{dacode,
  title={Da-code: Agent data science code generation benchmark for large language models},
  author={Huang, Yiming and Luo, Jianwen and Yu, Yan and Zhang, Yitong and Lei, Fangyu and Wei, Yifan and He, Shizhu and Huang, Lifu and Liu, Xiao and Zhao, Jun and Liu, Kang},
  booktitle={Proceedings of the 2024 Conference on Empirical Methods in Natural Language Processing},
  pages={13487--13521},
  year={2024}
}

@article{infiagent,
  title={InfiAgent-DABench: Evaluating Agents on Data Analysis Tasks},
  author={Hu, Xueyu and Zhao, Ziyu and Wei, Shuang and Chai, Ziwei and Ma, Qianli and Wang, Guoyin and Wang, Xuwu and Su, Jing and Xu, Jingjing and Zhu, Ming and Cheng, Yao and Yuan, Jianbo and Li, Jiwei and Kuang, Kun and Yang, yang and Yang, Hongxia and Wu, Fei},
  journal={Proceedings of Machine Learning Research},
  volume={235},
  pages={19544--19572},
  year={2024}
}

@article{fdabench,
  title={FDABench: A Benchmark for Data Agents on Analytical Queries over Heterogeneous Data},
  author={Wang, Ziting and Zhang, Shize and Yuan, Haitao and Zhu, Jinwei and Li, Shifu and Dong, Wei and Cong, Gao},
  journal={arXiv preprint arXiv:2509.02473},
  year={2025}
}

@article{dabstep,
  title={Dabstep: Data agent benchmark for multi-step reasoning},
  author={Egg, Alex and Goyanes, Martin Iglesias and Kingma, Friso and Mora, Andreu and von Werra, Leandro and Wolf, Thomas},
  journal={arXiv preprint arXiv:2506.23719},
  year={2025}
}

@inproceedings{feb4rag,
author = {Wang, Shuai and Khramtsova, Ekaterina and Zhuang, Shengyao and Zuccon, Guido},
title = {FeB4RAG: Evaluating Federated Search in the Context of Retrieval Augmented Generation},
year = {2024},
isbn = {9798400704314},
publisher = {Association for Computing Machinery},
address = {New York, NY, USA},
url = {https://doi.org/10.1145/3626772.3657853},
doi = {10.1145/3626772.3657853},
booktitle = {Proceedings of the 47th International ACM SIGIR Conference on Research and Development in Information Retrieval},
pages = {763–773},
numpages = {11},
location = {Washington DC, USA},
series = {SIGIR '24}
}

@inproceedings{webarena,
 author = {Zhou, Shuyan and Xu, Frank F and Zhu, Hao and Zhou, Xuhui and Lo, Robert and Sridhar, Abishek and Cheng, Xianyi and Ou, Tianyue and Bisk, Yonatan and Fried, Daniel and Alon, Uri and Neubig, Graham},
 booktitle = {International Conference on Learning Representations},
 editor = {B. Kim and Y. Yue and S. Chaudhuri and K. Fragkiadaki and M. Khan and Y. Sun},
 pages = {15585--15606},
 title = {WebArena: A Realistic Web Environment for Building Autonomous Agents},
 url = {https://proceedings.iclr.cc/paper_files/paper/2024/file/4410c0711e9154a7a2d26f9b3816d1ef-Paper-Conference.pdf},
 volume = {2024},
 year = {2024}
}

@article{swe-ci,
  title={SWE-CI: Evaluating agent capabilities in maintaining codebases via continuous integration},
  author={Chen, Jialong and Xu, Xander and Wei, Hu and Chen, Chuan and Zhao, Bing},
  journal={arXiv preprint arXiv:2603.03823},
  year={2026}
}

@inproceedings{agentboard,
 author = {Ma, Chang and Zhang, Junlei and Zhu, Zhihao and Yang, Cheng and Yang, Yujiu and Jin, Yaohui and Lan, Zhenzhong and Kong, Lingpeng and He, Junxian},
 booktitle = {Advances in Neural Information Processing Systems},
 doi = {10.52202/079017-2365},
 editor = {A. Globerson and L. Mackey and D. Belgrave and A. Fan and U. Paquet and J. Tomczak and C. Zhang},
 pages = {74325--74362},
 publisher = {Curran Associates, Inc.},
 title = {AgentBoard: An Analytical Evaluation Board of Multi-turn LLM Agents},
 url = {https://proceedings.neurips.cc/paper_files/paper/2024/file/877b40688e330a0e2a3fc24084208dfa-Paper-Datasets_and_Benchmarks_Track.pdf},
 volume = {37},
 year = {2024}
}

@misc{galileo,
  author       = {Galileo},
  title        = {Introducing Agentic Evaluations},
  year         = {2025},
  url          = {https://www.galileo.ai/blog/introducing-agentic-evaluations},
}

@article{deepresearchqa,
  title={DeepSearchQA: Bridging the Comprehensiveness Gap for Deep Research Agents},
  author={Gupta, Nikita and Chatterjee, Riju and Haas, Lukas and Tao, Connie and Wang, Andrew and Liu, Chang and Oiwa, Hidekazu and Gribovskaya, Elena and Ackermann, Jan and Blitzer, John and Goldshtein, Sasha and Das, Dipanjan},
  journal={arXiv preprint arXiv:2601.20975},
  year={2026}
}

@inproceedings{zhang2024OpenPRMBuildingOpendomain,
   author = {Zhang, Kaiyan and Zhang, Jiayuan and Li, Haoxin and Zhu, Xuekai and Hua, Ermo and Lv, Xingtai and Ding, Ning and Qi, Biqing and Zhou, Bowen},
 booktitle = {International Conference on Learning Representations},
 editor = {Y. Yue and A. Garg and N. Peng and F. Sha and R. Yu},
 pages = {80846--80868},
 title = {OpenPRM: Building Open-domain Process-based Reward Models with Preference Trees},
 url = {https://proceedings.iclr.cc/paper_files/paper/2025/file/c919a2b5ec1de69f2629f9119676e336-Paper-Conference.pdf},
 volume = {2025},
 year = {2025}
}

@inproceedings{yao2023react,
  title         = {{ReAct}: Synergizing Reasoning and Acting in Language Models},
  author        = {Yao, Shunyu and Zhao, Jeffrey and Yu, Dian and Du, Nan and Shafran, Izhak and Narasimhan, Karthik and Cao, Yuan},
  booktitle     = {The Eleventh International Conference on Learning Representations},
  year          = {2023},
  url           = {https://iclr.cc/virtual/2023/oral/12647},
  eprint        = {2210.03629},
  archiveprefix = {arXiv}
}

@inproceedings{spider2,
 author = {Lei, Fangyu and Chen, Jixuan and Ye, Yuxiao and Cao, Ruisheng and Shin, Dongchan and SU, Hongjin and Suo, Zhaoqing and Gao, Hongcheng and Hu, Wenjing and Yin, Pengcheng and Zhong, Victor and Xiong, Caiming and Sun, Ruoxi and Liu, Qian and Wang, Sida and Yu, Tao},
 booktitle = {International Conference on Learning Representations},
 editor = {Y. Yue and A. Garg and N. Peng and F. Sha and R. Yu},
 pages = {28691--28735},
 title = {Spider 2.0: Evaluating Language Models on Real-World Enterprise Text-to-SQL Workflows},
 url = {https://proceedings.iclr.cc/paper_files/paper/2025/file/46c10f6c8ea5aa6f267bcdabcb123f97-Paper-Conference.pdf},
 volume = {2025},
 year = {2025}
}

@inproceedings{jing2024dsbench,
 author = {Jing, Liqiang and Huang, Zhehui and Wang, Xiaoyang and Yao, Wenlin and Yu, Wenhao and Ma, Kaixin and Zhang, Hongming and Du, Xinya and Yu, Dong},
 booktitle = {International Conference on Learning Representations},
 editor = {Y. Yue and A. Garg and N. Peng and F. Sha and R. Yu},
 pages = {32597--32649},
 title = {DSBench: How Far Are Data Science Agents from Becoming Data Science Experts?},
 url = {https://proceedings.iclr.cc/paper_files/paper/2025/file/50e9ad960ae78b741a6b4fea533f2eaf-Paper-Conference.pdf},
 volume = {2025},
 year = {2025}
}

@inproceedings{lightman2023lets,
 author = {Lightman, Hunter and Kosaraju, Vineet and Burda, Yuri and Edwards, Harrison and Baker, Bowen and Lee, Teddy and Leike, Jan and Schulman, John  and Sutskever, Ilya and Cobbe, Karl},
 booktitle = {International Conference on Learning Representations},
 editor = {B. Kim and Y. Yue and S. Chaudhuri and K. Fragkiadaki and M. Khan and Y. Sun},
 pages = {39578--39601},
 title = {Let\textquotesingle s Verify Step by Step},
 url = {https://proceedings.iclr.cc/paper_files/paper/2024/file/aca97732e30bcf1303bc22ac3924fd16-Paper-Conference.pdf},
 volume = {2024},
 year = {2024}
}

@inproceedings{ouyang2022training,
  title         = {Training Language Models to Follow Instructions with Human Feedback},
  author        = {Ouyang, Long and Wu, Jeffrey and Jiang, Xu and Almeida, Diogo and Wainwright, Carroll and Mishkin, Pamela and Zhang, Chong and Agarwal, Sandhini and Slama, Katarina and Ray, Alex and Schulman, John and Hilton, Jacob and Kelton, Fraser and Miller, Luke and Simens, Maddie and Askell, Amanda and Welinder, Peter and Christiano, Paul F. and Leike, Jan and Lowe, Ryan},
  booktitle     = {Advances in Neural Information Processing Systems},
  volume        = {35},
  pages         = {27730--27744},
  year          = {2022},
  publisher     = {Curran Associates, Inc.},
  url           = {https://proceedings.neurips.cc/paper_files/paper/2022/hash/b1efde53be364a73914f58805a001731-Abstract.html},
  eprint        = {2203.02155},
  archiveprefix = {arXiv}
}

@article{cobbe2021training,
  title         = {Training Verifiers to Solve Math Word Problems},
  author        = {Cobbe, Karl and Kosaraju, Vineet and Bavarian, Mohammad and Chen, Mark and Jun, Heewoo and Kaiser, Lukasz and Plappert, Matthias and Tworek, Jerry and Hilton, Jacob and Nakano, Reiichiro and Hesse, Christopher and Schulman, John},
  journal       = {arXiv preprint arXiv:2110.14168},
  year          = {2021},
  doi           = {10.48550/arXiv.2110.14168},
  url           = {https://arxiv.org/abs/2110.14168},
  eprint        = {2110.14168},
  archiveprefix = {arXiv}
}

@inproceedings{agentbench,
 author = {Liu, Xiao and Yu, Hao and Zhang, Hanchen and Xu, Yifan and Lei, Xuanyu and Lai, Hanyu and Gu, Yu and Ding, Hangliang and Men, Kaiwen and Yang, Kejuan and Zhang, Shudan and Deng, Xiang and Zeng, Aohan and Du, Zhengxiao and Zhang, Chenhui and Shen, Sheng and Zhang, Tianjun and Su, Yu and Sun, Huan and Huang, Minlie and Dong, Yuxiao and Tang, Jie},
 booktitle = {International Conference on Learning Representations},
 editor = {B. Kim and Y. Yue and S. Chaudhuri and K. Fragkiadaki and M. Khan and Y. Sun},
 pages = {52989--53046},
 title = {AgentBench: Evaluating LLMs as Agents},
 url = {https://proceedings.iclr.cc/paper_files/paper/2024/file/e9df36b21ff4ee211a8b71ee8b7e9f57-Paper-Conference.pdf},
 volume = {2024},
 year = {2024}
}

@inproceedings{wang2024mathshepherd,
    title = "Math-Shepherd: Verify and Reinforce {LLM}s Step-by-step without Human Annotations",
    author = "Wang, Peiyi  and
      Li, Lei  and
      Shao, Zhihong  and
      Xu, Runxin  and
      Dai, Damai  and
      Li, Yifei  and
      Chen, Deli  and
      Wu, Yu  and
      Sui, Zhifang",
    editor = "Ku, Lun-Wei  and
      Martins, Andre  and
      Srikumar, Vivek",
    booktitle = "Proceedings of the 62nd Annual Meeting of the Association for Computational Linguistics (Volume 1: Long Papers)",
    month = aug,
    year = "2024",
    address = "Bangkok, Thailand",
    publisher = "Association for Computational Linguistics",
    url = "https://aclanthology.org/2024.acl-long.510/",
    doi = "10.18653/v1/2024.acl-long.510",
    pages = "9426--9439",
}

@inproceedings{wang2024codeact,
  title     = {Executable Code Actions Elicit Better {LLM} Agents},
  author    = {Wang, Xingyao and Chen, Yangyi and Yuan, Lifan and Zhang, Yizhe and Li, Yunzhu and Peng, Hao and Ji, Heng},
  booktitle = {Proceedings of the 41st International Conference on Machine Learning},
  series    = {Proceedings of Machine Learning Research},
  volume    = {235},
  pages     = {50208--50232},
  year      = {2024},
  publisher = {PMLR},
  url       = {https://proceedings.mlr.press/v235/wang24h.html},
  eprint    = {2402.01030},
  archiveprefix = {arXiv}
}

@article{zhong2025comprehensive,
  title         = {A Comprehensive Survey of Reward Models: Taxonomy, Applications, Challenges, and Future},
  author        = {Zhong, Jialun and Shen, Wei and Li, Yanzeng and Gao, Songyang and Lu, Hua and Chen, Yicheng and Zhang, Yang and Zhou, Wei and Gu, Jinjie and Zou, Lei},
  journal       = {arXiv preprint arXiv:2504.12328},
  year          = {2025},
  doi           = {10.48550/arXiv.2504.12328},
  url           = {https://arxiv.org/abs/2504.12328},
  eprint        = {2504.12328},
  archiveprefix = {arXiv}
}

@article{zheng2025surveyprm,
  title         = {A Survey of Process Reward Models: From Outcome Signals to Process Supervisions for Large Language Models},
  author        = {Zheng, Congming and Zhu, Jiachen and Ou, Zhuoying and Chen, Yuxiang and Zhang, Kangning and Shan, Rong and Zheng, Zeyu and Yang, Mengyue and Lin, Jianghao and Yu, Yong and Zhang, Weinan},
  journal       = {arXiv preprint arXiv:2510.08049},
  year          = {2025},
  doi           = {10.48550/arXiv.2510.08049},
  url           = {https://arxiv.org/abs/2510.08049},
  eprint        = {2510.08049},
  archiveprefix = {arXiv}
}

@inproceedings{setlur2025rewarding,
 author = {Setlur, Amrith and Nagpal, Chirag and Fisch, Adam and Geng, Xinyang and Eisenstein, Jacob and Agarwal, Rishabh and Agarwal, Alekh and Berant, Jonathan  and Kumar, Aviral},
 booktitle = {International Conference on Learning Representations},
 editor = {Y. Yue and A. Garg and N. Peng and F. Sha and R. Yu},
 pages = {60808--60838},
 title = {Rewarding Progress: Scaling Automated Process Verifiers for LLM Reasoning},
 url = {https://proceedings.iclr.cc/paper_files/paper/2025/file/98711dea460bdefe0e651ca23ec98ba2-Paper-Conference.pdf},
 volume = {2025},
 year = {2025}
}

@inproceedings{zhang2025lessons,
  title     = {The Lessons of Developing Process Reward Models in Mathematical Reasoning},
  author    = {Zhang, Zhenru and Zheng, Chujie and Wu, Yangzhen and Zhang, Beichen and Lin, Runji and Yu, Bowen and Liu, Dayiheng and Zhou, Jingren and Lin, Junyang},
  booktitle = {Findings of the Association for Computational Linguistics: ACL 2025},
  pages     = {10495--10516},
  year      = {2025},
  month     = jul,
  address   = {Vienna, Austria},
  publisher = {Association for Computational Linguistics},
  doi       = {10.18653/v1/2025.findings-acl.547},
  url       = {https://aclanthology.org/2025.findings-acl.547/}
}

@inproceedings{xi2026agentprm,
  title     = {{AgentPRM}: Process Reward Models for {LLM} Agents via Step-Wise Promise and Progress},
  author    = {Xi, Zhiheng and Liao, Chenyang and Li, Guanyu and Zhang, Zhihao and Chen, Wenxiang and Wang, Binghai and Jin, Senjie and Zhou, Yuhao and Guan, Jian and Wu, Wei and Ji, Tao and Gui, Tao and Zhang, Qi and Huang, Xuanjing},
  booktitle = {Proceedings of the ACM Web Conference 2026},
  pages     = {4184--4195},
  year      = {2026},
  publisher = {ACM},
  doi       = {10.1145/3774904.3792551},
  url       = {https://doi.org/10.1145/3774904.3792551}
}

@inproceedings{liu2026agentic,
  title         = {Agentic Reinforcement Learning with Implicit Step Rewards},
  author        = {Liu, Xiaoqian and Wang, Ke and Wu, Yuchuan and Huang, Fei and Li, Yongbin and Jiao, Jianbin and Zhang, Junge},
  booktitle     = {The Fourteenth International Conference on Learning Representations},
  year          = {2026},
  url           = {https://openreview.net/forum?id=ooROvpmxMV},
  eprint        = {2509.19199},
  archiveprefix = {arXiv}
}

@article{kandel2012enterprise,
  title   = {Enterprise Data Analysis and Visualization: An Interview Study},
  author  = {Kandel, Sean and Paepcke, Andreas and Hellerstein, Joseph M. and Heer, Jeffrey},
  journal = {IEEE Transactions on Visualization and Computer Graphics},
  volume  = {18},
  number  = {12},
  pages   = {2917--2926},
  year    = {2012},
  doi     = {10.1109/TVCG.2012.219}
}

@article{crisan2021baton,
  title   = {Passing the Data Baton: A Retrospective Analysis on Data Science Work and Workers},
  author  = {Crisan, Anamaria and Fiore-Gartland, Brittany and Tory, Melanie},
  journal = {IEEE Transactions on Visualization and Computer Graphics},
  volume  = {27},
  number  = {2},
  pages   = {1860--1870},
  year    = {2021},
  doi     = {10.1109/TVCG.2020.3030340}
}

@techreport{mckinsey2016analytics,
  title       = {The Age of Analytics: Competing in a Data-Driven World},
  author      = {Henke, Nicolaus and Bughin, Jacques and Chui, Michael and Manyika, James and Saleh, Tamim and Wiseman, Bill and Sethupathy, Guru},
  institution = {McKinsey Global Institute Research},
  year        = {2016},
  url         = {https://www.mckinsey.com/capabilities/quantumblack/our-insights/the-age-of-analytics-competing-in-a-data-driven-world}
}

@book{oecd2015ddi,
  author    = {{OECD}},
  title     = {Data-Driven Innovation: Big Data for Growth and Well-Being},
  publisher = {OECD Publishing},
  address   = {Paris},
  year      = {2015},
  doi       = {10.1787/9789264229358-en}
}

@inproceedings{chu2016cleaning,
  title     = {Data Cleaning: Overview and Emerging Challenges},
  author    = {Chu, Xu and Ilyas, Ihab F. and Krishnan, Sanjay and Wang, Jiannan},
  booktitle = {Proceedings of the 2016 International Conference on Management of Data (SIGMOD)},
  pages     = {2201--2206},
  year      = {2016},
  publisher = {ACM},
  doi       = {10.1145/2882903.2912574}
}

@article{hellerstein2018selfservice,
  title   = {Self-Service Data Preparation: Research to Practice},
  author  = {Hellerstein, Joseph M. and Heer, Jeffrey and Kandel, Sean},
  journal = {IEEE Data Engineering Bulletin},
  volume  = {41},
  number  = {2},
  pages   = {23--34},
  year    = {2018},
  url     = {http://sites.computer.org/debull/A18june/p23.pdf}
}

@misc{tianchi2023ashare,
  title        = {A-share Listed Companies Quarterly Revenue Prediction Dataset},
  author       = {{Alibaba Tianchi}},
  year         = {2023},
  howpublished = {\url{https://tianchi.aliyun.com/dataset/1074}},
  note         = {Licensed under CC BY-NC-SA 4.0}
}

\appendix
\etocdepthtag.toc{appendix} 
\twocolumn[%
  \section*{Additional Details}
  \label{sec:appendix}

  \etocsettocstyle{}{\par\addvspace{\dbltextfloatsep}}
  \etocsettagdepth{mainbody}{none}
  \etocsettagdepth{appendix}{subsubsection}
  \tableofcontents
]


\section{Experimental Details}
\label{appendix:exp-details}

For every evaluated model we call its native vendor API with default setting \texttt{temperature}, \texttt{top\_p}, or \texttt{max\_tokens}. 
We likewise use each vendor's default reasoning or thinking behavior, summarised in Table~\ref{tab:model-configs}. 
Each agent is allocated its model's maximum supported context window, the OpenClaw orchestration layer does not impose an additional setting. 
The OpenClaw framework, Docker images, judge prompt, and per-task wall-clock timeout (1200\,s) are kept identical across models and difficulty buckets, as stated in \S\ref{sec:experiments}.

\begin{table}[!htbp]
\centering
\small
\setlength{\tabcolsep}{5pt}
\begin{tabular}{lcc}
\toprule
\renewcommand{\arraystretch}{1.1}
\textbf{Model} & \textbf{Context} & \textbf{Thinking Mode} \\
\midrule
Claude Opus 4.6        & 1M    & high effort            \\
Gemini 3.1 Pro Preview & 1M    & high effort                        \\
GPT-5.4                & 1.05M & off \\
DeepSeek-V3.2          & 131K  & thinking  \\
Minimax M2.7           & 197K  & thinking            \\
Qwen3.5-Plus           & 1M    & thinking                          \\
GLM-5                  & 203K  & thinking                          \\
Kimi-K2.5              & 262K  & thinking                          \\
\bottomrule
\end{tabular}
\caption{model evaluation configuration. ``Thinking Mode'' lists each model's vendor-API default for its reasoning / extended-thinking mode at the time of evaluation.}
\label{tab:model-configs}
\end{table}


\section{Supplementary Results}
\label{appendix:supplementary}

\subsection{Cross-Task and Cross-Difficulty Results}
\label{appendix:cross-task}

\begin{figure}[!htbp]
\centering
\includegraphics[width=0.9\linewidth]{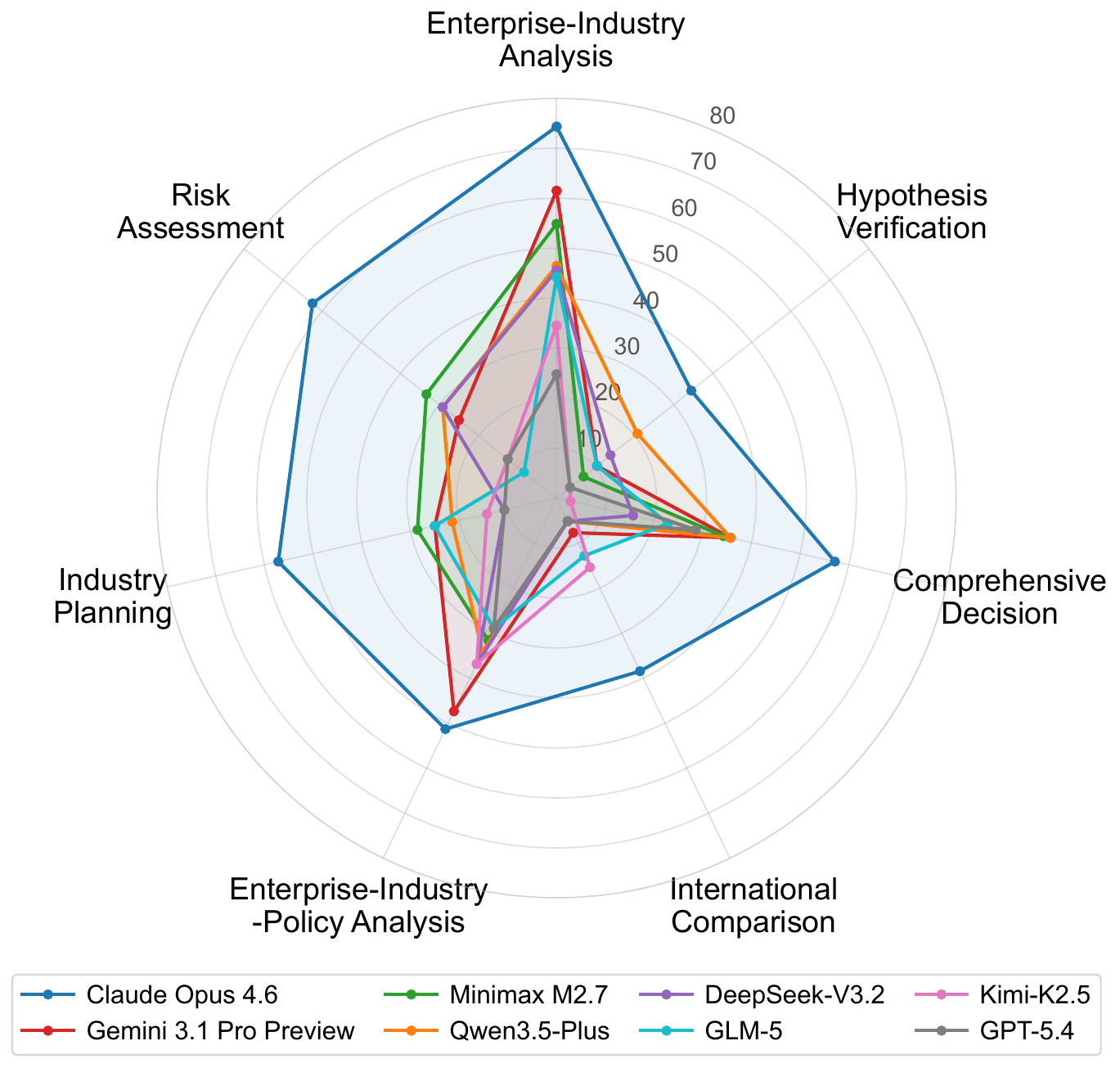}
\caption{Accuracy by task category across all models.}
\label{fig:difficulty-curve}
\end{figure}

Figure~\ref{fig:difficulty-curve} and Tables~\ref{tab:cross-task-claude}--\ref{tab:cross-task-kimi} report per-model accuracy broken down by task category and difficulty level. Rows correspond to task categories and columns to difficulty levels. A ``--'' entry indicates that the corresponding category has no tasks at the given difficulty.

\begin{table*}[!htbp]
\centering
\small
\renewcommand{\arraystretch}{1.1}
\begin{tabular}{lcccc}
\toprule
\textbf{Category} & \textbf{Easy} & \textbf{Medium} & \textbf{Hard} & \textbf{Overall} \\
\midrule
Enterprise--Industry Analysis & 76.5 & 72.1 & -- & 74.3 \\
Enterprise--Industry--Policy Analysis & 80.0 & 47.0 & -- & 51.3 \\
Comprehensive Decision & 76.7 & 65.6 & 36.8 & 58.7 \\
International Comparison & -- & 52.3 & 50.0 & 51.5 \\
Hypothesis Verification & -- & 50.0 & 26.1 & 37.6 \\
Industry Planning & -- & 71.4 & 54.8 & 63.1 \\
Risk Assessment & -- & 66.7 & 61.5 & 63.9 \\
\midrule
Overall & 76.8 & 62.2 & 44.8 & 63.4 \\
\bottomrule
\end{tabular}
\caption{Accuracy (\%) of Claude Opus 4.6 by task category and difficulty.}
\label{tab:cross-task-claude}
\end{table*}

\begin{table*}[!htbp]
\centering
\small
\renewcommand{\arraystretch}{1.1}
\begin{tabular}{lcccc}
\toprule
\textbf{Category} & \textbf{Easy} & \textbf{Medium} & \textbf{Hard} & \textbf{Overall} \\
\midrule
Enterprise--Industry Analysis & 43.5 & 49.5 & -- & 46.5 \\
Enterprise--Industry--Policy Analysis & 70.0 & 28.8 & -- & 34.2 \\
Comprehensive Decision & 33.3 & 50.0 & 16.7 & 39.5 \\
International Comparison & -- & 15.3 & 17.9 & 16.2 \\
Hypothesis Verification & -- & 42.9 & 0.0 & 20.7 \\
Industry Planning & -- & 35.7 & 10.7 & 23.2 \\
Risk Assessment & -- & 27.3 & 38.5 & 33.3 \\
\midrule
Overall & 45.0 & 40.0 & 16.2 & 37.7 \\
\bottomrule
\end{tabular}
\caption{Accuracy (\%) of Qwen3.5-Plus by task category and difficulty.}
\label{tab:cross-task-qwen35plus}
\end{table*}

\begin{table*}[!htbp]
\centering
\small
\renewcommand{\arraystretch}{1.1}
\begin{tabular}{lcccc}
\toprule
\textbf{Category} & \textbf{Easy} & \textbf{Medium} & \textbf{Hard} & \textbf{Overall} \\
\midrule
Enterprise--Industry Analysis & 45.2 & 45.9 & -- & 45.6 \\
Enterprise--Industry--Policy Analysis & 60.0 & 33.3 & -- & 36.8 \\
Comprehensive Decision & 16.7 & 28.4 & 3.5 & 20.6 \\
International Comparison & -- & 22.3 & 14.3 & 19.4 \\
Hypothesis Verification & -- & 28.6 & 0.0 & 13.8 \\
Industry Planning & -- & 21.4 & 0.0 & 10.7 \\
Risk Assessment & -- & 36.4 & 30.8 & 33.3 \\
\midrule
Overall & 45.0 & 35.8 & 8.9 & 34.1 \\
\bottomrule
\end{tabular}
\caption{Accuracy (\%) of DeepSeek-V3.2 by task category and difficulty.}
\label{tab:cross-task-deepseek}
\end{table*}

\begin{table*}[!htbp]
\centering
\small
\renewcommand{\arraystretch}{1.1}
\begin{tabular}{lcccc}
\toprule
\textbf{Category} & \textbf{Easy} & \textbf{Medium} & \textbf{Hard} & \textbf{Overall} \\
\midrule
Enterprise--Industry Analysis & 43.5 & 45.0 & -- & 44.2 \\
Enterprise--Industry--Policy Analysis & 80.0 & 21.2 & -- & 28.9 \\
Comprehensive Decision & 33.3 & 28.9 & 5.3 & 22.9 \\
International Comparison & -- & 35.6 & 28.0 & 32.9 \\
Hypothesis Verification & -- & 14.3 & 6.7 & 10.3 \\
Industry Planning & -- & 35.7 & 17.9 & 26.8 \\
Risk Assessment & -- & 13.6 & 7.7 & 10.4 \\
\midrule
Overall & 45.8 & 33.0 & 12.6 & 33.3 \\
\bottomrule
\end{tabular}
\caption{Accuracy (\%) of GLM-5 by task category and difficulty.}
\label{tab:cross-task-glm5}
\end{table*}

\begin{table*}[!htbp]
\centering
\small
\renewcommand{\arraystretch}{1.1}
\begin{tabular}{lcccc}
\toprule
\textbf{Category} & \textbf{Easy} & \textbf{Medium} & \textbf{Hard} & \textbf{Overall} \\
\midrule
Enterprise--Industry Analysis & 19.1 & 30.6 & -- & 24.8 \\
Enterprise--Industry--Policy Analysis & 80.0 & 21.2 & -- & 28.9 \\
Comprehensive Decision & 3.3 & 44.4 & 8.8 & 31.2 \\
International Comparison & -- & 17.3 & 14.3 & 16.2 \\
Hypothesis Verification & -- & 7.1 & 2.2 & 4.6 \\
Industry Planning & -- & 10.7 & 14.3 & 12.5 \\
Risk Assessment & -- & 9.1 & 23.1 & 16.7 \\
\midrule
Overall & 23.1 & 26.5 & 12.0 & 23.4 \\
\bottomrule
\end{tabular}
\caption{Accuracy (\%) of GPT-5.4 by task category and difficulty.}
\label{tab:cross-task-gpt54}
\end{table*}

\begin{table*}[!htbp]
\centering
\small
\renewcommand{\arraystretch}{1.1}
\begin{tabular}{lcccc}
\toprule
\textbf{Category} & \textbf{Easy} & \textbf{Medium} & \textbf{Hard} & \textbf{Overall} \\
\midrule
Enterprise--Industry Analysis & 67.0 & 55.9 & -- & 61.5 \\
Enterprise--Industry--Policy Analysis & 80.0 & 42.4 & -- & 47.4 \\
Comprehensive Decision & 60.0 & 48.4 & 11.8 & 39.5 \\
International Comparison & -- & 14.0 & 3.6 & 10.3 \\
Hypothesis Verification & -- & 21.4 & 0.0 & 10.3 \\
Industry Planning & -- & 42.9 & 10.7 & 26.8 \\
Risk Assessment & -- & 28.8 & 38.5 & 34.0 \\
\midrule
Overall & 67.6 & 44.6 & 12.3 & 45.8 \\
\bottomrule
\end{tabular}
\caption{Accuracy (\%) of Gemini 3.1 Pro Preview by task category and difficulty.}
\label{tab:cross-task-gemini}
\end{table*}

\begin{table*}[!htbp]
\centering
\small
\renewcommand{\arraystretch}{1.1}
\begin{tabular}{lcccc}
\toprule
\textbf{Category} & \textbf{Easy} & \textbf{Medium} & \textbf{Hard} & \textbf{Overall} \\
\midrule
Enterprise--Industry Analysis & 65.2 & 44.1 & -- & 54.9 \\
Enterprise--Industry--Policy Analysis & 40.0 & 30.3 & -- & 31.6 \\
Comprehensive Decision & 50.0 & 45.6 & 16.7 & 38.1 \\
International Comparison & -- & 28.0 & 10.7 & 21.8 \\
Hypothesis Verification & -- & 14.3 & 0.0 & 6.9 \\
Industry Planning & -- & 39.3 & 25.0 & 32.1 \\
Risk Assessment & -- & 22.7 & 48.7 & 36.8 \\
\midrule
Overall & 62.6 & 37.2 & 19.3 & 41.3 \\
\bottomrule
\end{tabular}
\caption{Accuracy (\%) of Minimax M2.7 by task category and difficulty.}
\label{tab:cross-task-minimax}
\end{table*}

\begin{table*}[!htbp]
\centering
\small
\renewcommand{\arraystretch}{1.1}
\begin{tabular}{lcccc}
\toprule
\textbf{Category} & \textbf{Easy} & \textbf{Medium} & \textbf{Hard} & \textbf{Overall} \\
\midrule
Enterprise--Industry Analysis & 41.7 & 27.0 & -- & 34.5 \\
Enterprise--Industry--Policy Analysis & 60.0 & 33.3 & -- & 36.8 \\
Comprehensive Decision & 0.0 & 7.3 & 0.0 & 4.7 \\
International Comparison & -- & 22.7 & 32.1 & 26.1 \\
Hypothesis Verification & -- & 0.0 & 6.7 & 3.4 \\
Industry Planning & -- & 21.4 & 14.3 & 17.9 \\
Risk Assessment & -- & 7.6 & 23.1 & 16.0 \\
\midrule
Overall & 41.2 & 22.6 & 14.0 & 26.3 \\
\bottomrule
\end{tabular}
\caption{Accuracy (\%) of Kimi-K2.5 by task category and difficulty.}
\label{tab:cross-task-kimi}
\end{table*}

\subsection{Operation and Termination Taxonomies}
\label{appendix:operation-taxonomy}

Table~\ref{tab:op-by-term} cross-tabulates two orthogonal taxonomies: what operation the agent was attempting when it first deviated from the gold milestone chain, and how the run ultimately terminated. We define each taxonomy below.

\paragraph{Operation taxonomy.} We use a six-class operation taxonomy applied to the first un-achieved milestone of each failed run. The six classes are:
\begin{itemize}
    \item \textbf{Entity Attribute Lookup}, mapping a single named entity to a single attribute, for example ``Industry of \textit{X} Company'' or ``Operating revenue of \textit{X} in 2022''.
    \item \textbf{Aggregate Count or Sum}, counting or summing across a filtered group, for example ``Total chemical enterprises'' or ``Number of provinces with valid disclosure''.
    \item \textbf{Statistical Summary}, computing a median, mean, a derived ratio such as per-capita or year-over-year change, or a composite normalized score.
    \item \textbf{Policy Lookup and Count}, querying or counting policy records by theme $\times$ issuer $\times$ region.
    \item \textbf{Comparison or Boolean Judgment}, performing pairwise comparison, threshold judgment, difference, or yes-or-no conclusion.
    \item \textbf{Ranking and Selection}, returning a top-$k$ list, a rank position, or selection of the first, last, highest, or lowest item.
\end{itemize}

\paragraph{Termination taxonomy.}
Each failed run is assigned exactly one termination cause according to how the trajectory ends. 
We use five priority-ordered classes. 
\begin{itemize}
    \item \textbf{Timeout Kill} denotes runs externally terminated after exhausting the time budget. 
    \item \textbf{Silent Stop} denotes runs that end without a usable final response. This often occurs when transient context saturation causes the model to stop reasoning. 
    \item \textbf{Voluntary Give-up} denotes runs where the agent explicitly states that it cannot find the required evidence or that the requested information appears unavailable. 
    \item \textbf{Wrong-Answer Stop} denotes runs where the agent emits a non-empty final answer and stops as if the task were solved, but the answer is incorrect. 
    \item \textbf{Other} collects residual cases that do not fit the above categories.
\end{itemize}

\subsection{Environmental Impact Analysis}
\label{appendix:env-impact}

\begin{figure}[!htbp]
\centering
\includegraphics[width=\linewidth]{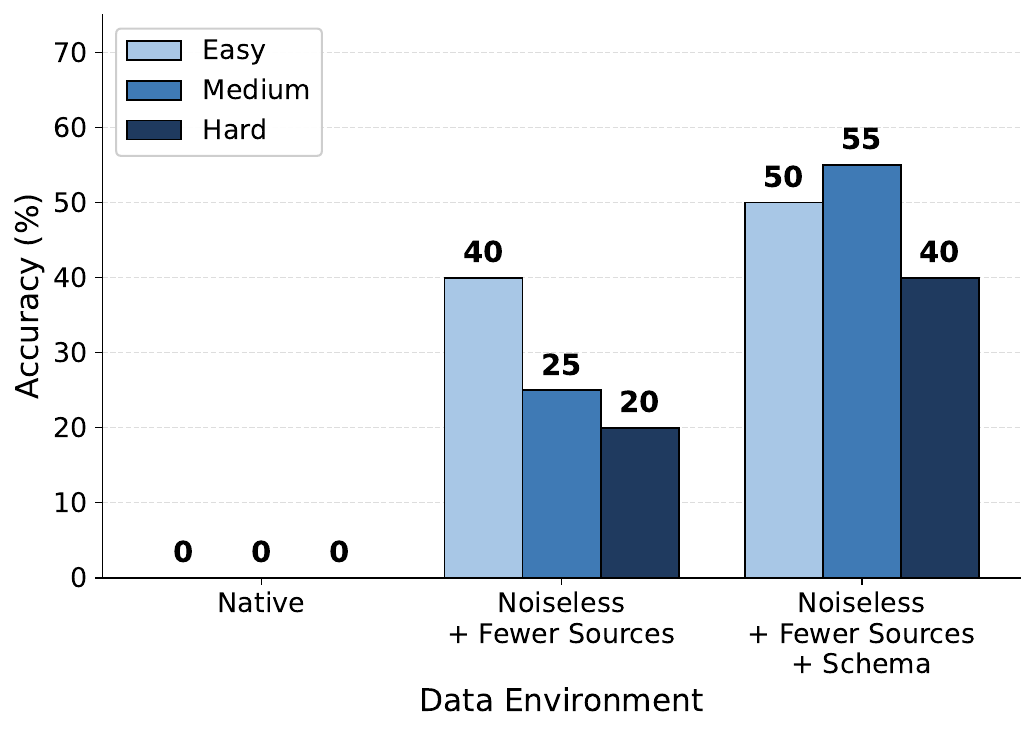}
\caption{GLM-5 accuracy under progressively cleaned data environments. }
\label{fig:glm5_model_accuracy_different_data_condition}
\end{figure}

To verify that the noise and reasoning decomposition reported in \S\ref{subsubsec:claim2} is not specific to Qwen3.5-Plus, we replicate the paired rerun study on a second model under the same three progressively cleaned data environments. Figure~\ref{fig:glm5_model_accuracy_different_data_condition} shows that GLM-5 reproduces the same directional pattern as Qwen3.5-Plus in Figure~\ref{fig:main-cleaning}, where cleaning improves Easy and Medium accuracy more than Hard accuracy. The phenomenon is therefore reproducible across models rather than an artifact of any single backbone.

\subsection{Cost Analysis}
\label{appendix:cost}

Table~\ref{tab:cost} reports the computational cost of evaluating the DataClawBench benchmark for each model, priced at OpenRouter list rates as of 2026-04-25 (full unit prices in Table~\ref{tab:pricing}). Per-task cost is the sum of input, output, cache-read, and cache-write token costs computed from each model's own per-million-token rates.

\paragraph{Cost patterns.}
The cost spread across models is governed by three distinct forces along the price-volume axis. At the high end, premium per-token rates dominate. Claude Opus 4.6 charges \$5 and \$25 per million input and output tokens with no cache-read reuse, accumulating roughly 40.0\% of the total benchmark spend. In the middle band, volume and rate trade off. Qwen3.5-Plus processes the most input tokens per task at 1.84M but lands mid-pack in total cost because its per-million input rate is roughly twenty times lower than Opus. At the low end, cache-heavy pathways and cheap rates take over. Minimax M2.7 stays cheapest at roughly \$0.12 per task, a cost gap of more than an order of magnitude below Opus, supported by the largest cache-read volume at \$0.059 per million; GLM-5 and Kimi-K2.5 follow with substantial cache-read volumes priced at \$0.12 and \$0.22 per million.

\paragraph{Per-outcome request counts and exploration archetypes.}
The per-outcome request columns ``Avg Reqs (C)'' and ``Avg Reqs (W)'' supply the request-count evidence behind the four archetypes introduced in \S\ref{subsubsec:efficiency}, walked through here in the same order. The decisive solver Claude Opus 4.6 issues only 12.5 requests per correctly answered task and 18.9 per failed task, the lowest combination among models with high GPR. The persistent-but-late group of Gemini 3.1 Pro Preview and Minimax M2.7 sits in a similar range on correct tasks at 19.7 and 21.7 but expands sharply on failed tasks to 41.8 and 32.1, consistent with extended exploration that arrives at milestones late rather than early. The wasteful trial-and-error group of DeepSeek-V3.2 and Qwen3.5-Plus runs high on both columns, at 30.5 and 24.9 on correct tasks and 36.6 and 32.8 on failed tasks, spending many steps without proportionate progress. The disengaged group of GPT-5.4, Kimi-K2.5, and GLM-5 sits at or below 19 on both columns, terminating early on tasks they cannot solve.

Contrasting the decisive solver with the persistent-but-late group sharpens the cost-versus-progress tradeoff. Claude reaches 45.1\% GPR with only 18.9 failed-task requests, whereas Gemini and Minimax issue 41.8 and 32.1 requests for lower GPR of 33.6\% and 33.2\%. Per failed-task request, Claude extracts roughly three times the milestone progress that Gemini does, mirroring Claude's substantially higher TPE of 0.59 against Gemini's 0.41 and Minimax's 0.37.

\begin{table*}[!htbp]
\centering
\small
\renewcommand{\arraystretch}{1.2}
\resizebox{\textwidth}{!}{%
\begin{tabular}{lrrrrrrrrrr}
\toprule
\textbf{Model} & \textbf{Tasks} & \textbf{Reqs} & \textbf{Total Tok} & \textbf{Avg Input} & \textbf{Avg Output} & \textbf{Avg Cache} & \textbf{Avg Reqs (C)} & \textbf{Avg Reqs (W)} & \textbf{Cost/Task} & \textbf{Total} \\
\midrule
Claude Opus 4.6        & 492 & 7{,}358  & 282M & 569K    & 4.5K  & 0     & 12.5 & 18.9 & \$2.956 & \$1{,}454.26 \\
Qwen3.5-Plus           & 492 & 14{,}739 & 907M & 1{,}840K & 6.5K  & 0     & 24.9 & 32.8 & \$0.488 &   \$240.01 \\
DeepSeek-V3.2          & 492 & 17{,}033 & 700M & 1{,}371K & 9.9K  & 43K   & 30.5 & 36.6 & \$0.350 &   \$172.17 \\
GLM-5                  & 492 & 9{,}046  & 408M & 173K    & 6.0K  & 654K  & 17.1 & 19.1 & \$0.194 &    \$95.37 \\
GPT-5.4                & 492 & 5{,}610  & 207M & 325K    & 2.3K  & 93K   &  9.9 & 11.8 & \$0.870 &   \$428.21 \\
Gemini 3.1 Pro Preview & 492 & 15{,}730 & 751M & 926K    & 9.9K  & 590K  & 19.7 & 41.8 & \$2.089 & \$1{,}027.96 \\
Minimax M2.7           & 492 & 13{,}806 & 553M & 182K    & 12.4K & 930K  & 21.7 & 32.1 & \$0.124 &    \$61.15 \\
Kimi-K2.5              & 492 & 7{,}883  & 465M & 517K    & 2.3K  & 427K  & 16.1 & 16.0 & \$0.326 &   \$160.25 \\
\midrule
\textbf{Total}         & 3{,}934 & 91{,}197 & 4{,}273M & -- & -- & -- & -- & -- & -- & \textbf{\$3{,}639.40} \\
\bottomrule
\end{tabular}%
}
\caption{Computational cost of evaluating DataClawBench, priced at OpenRouter list rates as of 2026-04-25. ``Reqs'' is the total number of model API requests across those tasks. ``Total Tok'' is the total token volume. Per-task columns show averages over tasks. ``Avg Reqs (C)'' and ``Avg Reqs (W)'' are the average number of requests per task on correctly and incorrectly answered tasks, respectively. ``Cost/Task'' is the sum of input, output, cache-read, and cache-write costs at each model's OpenRouter per-million-token rate (see Table~\ref{tab:pricing}).}
\label{tab:cost}
\end{table*}

\begin{table*}[!htbp]
\centering
\small
\renewcommand{\arraystretch}{1.1}
\begin{tabular}{lrrrr}
\toprule
\textbf{Model} & \textbf{Input} & \textbf{Output} & \textbf{Cache Read} & \textbf{Cache Write} \\
\midrule
Claude Opus 4.6        & \$5.00  & \$25.00 & \$0.50   & \$6.25  \\
Qwen3.5-Plus           & \$0.26  & \$1.56  & --       & \$0.325 \\
DeepSeek-V3.2          & \$0.252 & \$0.378 & \$0.0252 & --      \\
GLM-5                  & \$0.60  & \$2.08  & \$0.12   & --      \\
GPT-5.4                & \$2.50  & \$15.00 & \$0.25   & --      \\
Gemini 3.1 Pro Preview & \$2.00  & \$12.00 & \$0.20   & \$0.375 \\
Minimax M2.7           & \$0.30  & \$1.20  & \$0.059  & --      \\
Kimi-K2.5              & \$0.44  & \$2.00  & \$0.22   & --      \\
\bottomrule
\end{tabular}
\caption{Per-million-token unit prices on OpenRouter as of 2026-04-25 (USD). ``--'' indicates the price tier is not advertised by the provider.}
\label{tab:pricing}
\end{table*}

\subsection{Cross-Benchmark Accuracy Illustration}
\label{app:cross_benchmark}

\begin{figure}[!htbp]
\centering
\includegraphics[width=\columnwidth]{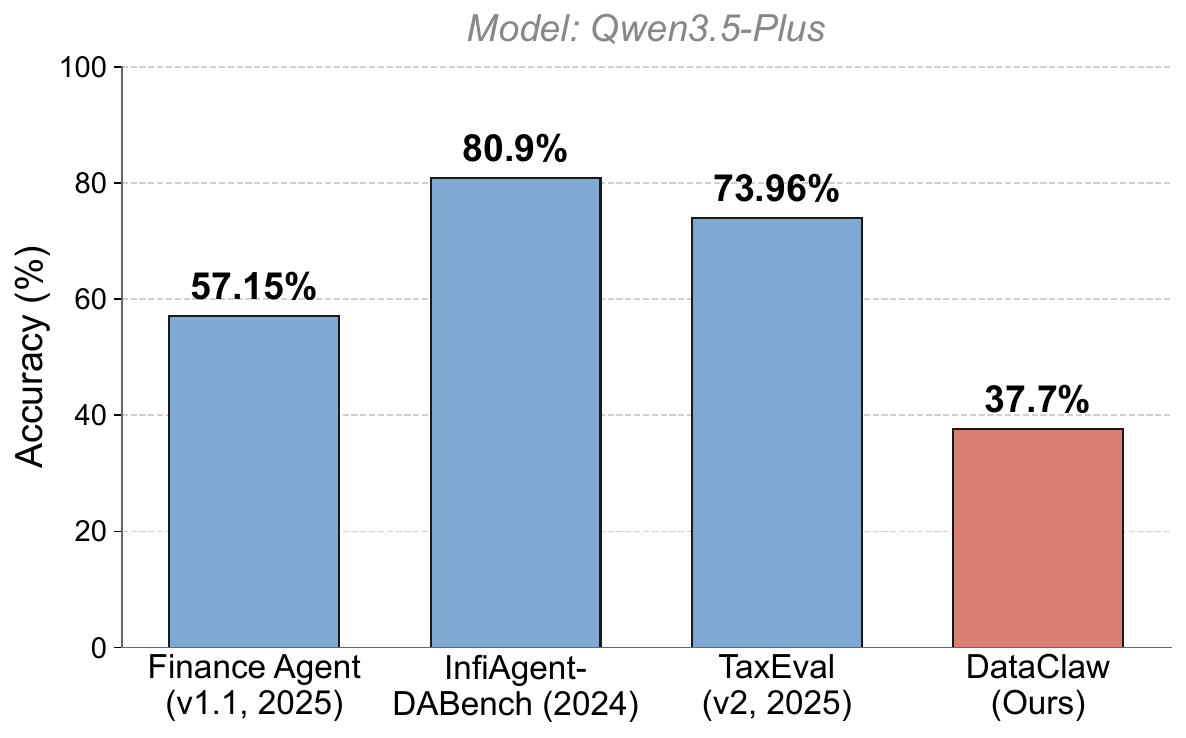}
\caption{Accuracy across data analysis benchmarks. }
\label{fig:benchmark-comparison}
\end{figure}

Figure~\ref{fig:benchmark-comparison} provides an illustrative comparison of Qwen3.5-Plus accuracy across several recent data-analysis benchmarks.
This comparison is not intended as a controlled ranking across benchmarks.
Different benchmarks vary in task design, evaluation protocol, tool environment, and result source.
Finance Agent and TaxEval results are from official leaderboards, while others are evaluated with OpenClaw. 
Therefore, the figure should be interpreted only as a coarse indication that DataClawBench poses a challenging evaluation setting.

\section{Detailed Comparison with Dense Process Reward Models}
\label{appendix:mrm-vs-prm}

Our results show that real-world data analysis depends not only on final-answer correctness but also on whether the execution trajectory makes sustained progress through critical intermediate states. Current agent training relies primarily on final-answer supervision, which is insufficient for the open-ended exploration and multi-stage analysis these tasks require~\cite{yao2023react,infiagent,jing2024dsbench}.

The milestone annotations and process scoring mechanism provided by DataClawBench make it possible to transform whether an agent reaches key intermediate states during the process into milestone-level reward signals~\cite{ouyang2022training}. Based on these signals, one can train a reward model that lies between a sparse Outcome Reward Model (ORM) and a dense Process Reward Model (PRM). Compared with ORM signals~\cite{zhong2025comprehensive}, these signals are denser and can help alleviate the supervision sparsity and credit assignment difficulties~\cite{cobbe2021training,lightman2023lets}.
Compared with PRM signals, they are coarser but naturally aligned with the goal-oriented and path-diverse nature of real-world workflows. Thus, DataClawBench enables progress-based verifier training and trajectory-level reward shaping in the data-analysis setting.

More specifically, compared with dense PRM, which provides stepwise rewards~\cite{zheng2025surveyprm}, milestone-based supervision can mitigate three issues. First, dense PRM often requires costly human annotation~\cite{lightman2023lets}, or relies on extensive sampling to construct training corpus~\cite{wang2024mathshepherd,zhang2024OpenPRMBuildingOpendomain}. Second, annotation methods that infer step-level values from final rewards are susceptible to sampling variance and to the quality of automatic annotation strategies~\cite{wang2024mathshepherd,zhang2024OpenPRMBuildingOpendomain,setlur2025rewarding,zhang2025lessons,liu2026agentic}. Third, exploratory retrieval and local backtracking are often necessary for information gathering and subsequent error correction. Dense PRM tends to mistakenly penalize such actions, whose short-term benefits may be unclear but which are indispensable for subsequent progress~\cite{wang2024codeact,xi2026agentprm,setlur2025rewarding}.

\section{Supplementary Data and Task Details}
\label{appendix:data-task}

\subsection{Data Environment Details}
\label{appendix:data-env}

The data environment is organized around three thematic domains of enterprise, industry, and policy. Subcategories cover enterprise profiles together with region-specific variants, enterprise core competitiveness, business status, regional and national industry statistics, policy releases, and full policy text. The taxonomy follows the structure of real research and consulting workflows. The environment contains 18 independent data files, where each file counts as one source. Of these, 17 reside under the three theme subdirectories, and one serves as an internal business-logic knowledge base unassigned to any theme domain.

\begin{table*}[!htbp]
\centering
\small
\renewcommand{\arraystretch}{1.1}
\begin{tabular}{l|c|p{10.5cm}}
\toprule
\textbf{Dimension} & \textbf{Value} & \textbf{Notes} \\
\midrule
Domains & 3 & Enterprise, industry, and policy. Counts only the 17 theme-domain files. \\
Core data sources & 7 & Three enterprise sources for profiles, core competitiveness, and business status. Two industry sources for regional and national statistics. Two policy sources for release status and full text. Per-file mapping appears in Table~\ref{tab:data-full}. \\
Total source files & 18 & Each file is treated as one source. All files are mounted into the container workspace at runtime. Composition is detailed in Table~\ref{tab:data-full}. \\
Format & Mainly CSV & Files are primarily CSV and contain both structured fields and long-form unstructured text. \\
Time span & Mainly 2022 & Statistical periods vary across sources. \\
Data usability issues & -- & Missing values, definition mismatches, inconsistent naming, and others, reflecting real-world data noise. \\
\bottomrule
\end{tabular}
\caption{Data Environment Summary}
\label{tab:data-environment-statistic}
\end{table*}

\begin{table*}[!htbp]
\centering
\small
\begin{tabular}{lllrrp{4cm}}
\toprule
\textbf{Domain} & \textbf{Source} & \textbf{Table} & \textbf{Records} & \textbf{Fields} & \textbf{Key Attributes} \\
\midrule
\multirow{9}{*}{Enterprise}
& \multirow{5}{*}{company profile}
& company\_profile           & 7,295   & 29 & bmCode, industry, province, listing date \\
& & company\_profile\_as     & 1,282   & 29 & Asia regional subset, identical schema \\
& & company\_profile\_eu     & 1,201   & 29 & Europe regional subset, identical schema \\
& & company\_profile\_na     & 791     & 29 & North America regional subset \\
& & company\_profile\_oc     & 401     & 29 & Oceania regional subset \\
\cmidrule(lr){2-6}
& \multirow{3}{*}{company operation}
& company\_operation\_status         & 353,438 & 9 & Annual KPIs covering assets, revenue, profit, and R\&D \\
& & company\_operation\_status\_* & 781,772 & 5 & Per-indicator detail rows for each enterprise-year \\
& & company\_operation\_yearly\_* & 669,169 & 5 & Year-over-year change metrics by enterprise \\
\cmidrule(lr){2-6}
& company core & company\_core & 4,907 & 8 & Core competitiveness narratives \\
\midrule
\multirow{6}{*}{Industry}
& \multirow{3}{*}{nation-wide industry}
& national\_industry\_status         & 12,107  & 9  & National industry aggregates \\
& & national\_industry\_status\_* & 43,938  & 7  & Granular per-indicator national breakdowns \\
& & national\_industry\_yearly\_* & 776     & 7  & Year-over-year national aggregates \\
\cmidrule(lr){2-6}
& \multirow{3}{*}{regional industry}
& regional\_industry\_status         & 170,896 & 10 & Provincial industry aggregates \\
& & regional\_industry\_status\_* & 3,173   & 8  & Granular per-indicator provincial breakdowns \\
& & regional\_industry\_yearly\_* & 3,207   & 8  & Year-over-year provincial aggregates \\
\midrule
\multirow{2}{*}{Policy}
& policy resource
& policy\_resource           & 1,129   & 19 & Full policy texts with metadata \\
\cmidrule(lr){2-6}
& policy release
& policy\_release\_status    & 7,792   & 9  & Policy publication statistics \\
\midrule
Internal & -- & internal\_metrics & 25 & 2 & Metric definitions (EN + ZH) \\
\bottomrule
\end{tabular}
\caption{Complete data environment inventory. \textbf{Domain} denotes the top-level analytical dimension under the Enterprise--Industry--Policy framework. \textbf{Source} indicates the thematic data source that groups related tables within a domain (e.g., \textit{company profile} aggregates the global table and its four regional subsets). \textbf{Table} is the physical CSV/relational table name on disk. \textbf{Records} reports the number of rows. \textbf{Fields} reports the number of columns in the table schema. \textbf{Key Attributes} summarizes the representative columns or semantic role of each table.}
\label{tab:data-full}
\end{table*}

Table~\ref{tab:data-full} provides a complete inventory of all 18 tables in the DataClawBench data environment, including field counts, record counts, and representative fields. The majority of these tables are independently collected and curated by the author team, while a subset of the Enterprise and Industry tables additionally incorporates open-source curated financial data~\cite{tianchi2023ashare}.

\subsection{Data Anonymization Details}
\label{appendix:anonymization}

To preserve the dataset's inferential value and cross-domain relational structure while protecting commercial privacy and meeting academic compliance requirements, DataClawBench applies a three-stage de-identification pipeline of identifier pseudonymization, distribution-preserving perturbation, and cross-domain consistency validation.

\paragraph{Identifier Pseudonymization.} Key identifiers capable of directly revealing corporate identity are irreversibly pseudonymized. Stock codes are mapped to pseudo-identifiers through custom cryptographic hashing, removing any deterministic linkage to the original securities. Enterprise names are reconstructed under controlled stochastic substitution that respects industry-specific semantic features and conventional naming patterns, retaining contextual plausibility within each industry while blocking reverse identification.
\paragraph{Distribution-Preserving Perturbation.} Quantitative operational metrics such as revenue, profit, R\&D expenditure, and patent counts are perturbed with a least-significant-digit mechanism. The mechanism applies controlled random perturbation only to the low-order digits of each value, preserving overall magnitude and relative rankings while disrupting exact-match identification. Unlike global additive noise, the perturbation targets only the low-precision representation layer of the data, retaining the original statistical moments such as mean, variance, skewness, and kurtosis as well as industry quantile structure. Kolmogorov--Smirnov tests confirm that the cumulative distribution functions of sub-industry subsets show no significant deviation between pre- and post-perturbation samples ($p > 0.05$), preserving the statistical fidelity required for cross-enterprise comparison, trend extrapolation, and econometric modelling.
\paragraph{Cross-Domain Consistency Validation.} To prevent perturbation-induced inconsistencies in industry-level aggregates, we re-derive industry aggregate indicators from the perturbed enterprise data and align them with the original industry-level statistics on aggregate structure, mean offset, and distribution shape. The procedure preserves the data noise inherent in real data ecosystems, including mismatches in statistical scope, financial rounding conventions, cross-period disclosure lags, and measurement errors. As a result, the de-identified dataset retains the relational structure and causal constraints required for cross-domain inference and knowledge graph construction.

Together, these stages eliminate individual re-identification risk while preserving the structural, statistical, and business-logic associations across domains required for analytical use.

\subsection{Task Construction Details}
\label{appendix:generation}

\subsubsection{Data Sources and De-identification Preprocessing}

The raw corpus of DataClawBench is not synthesized by large language models nor drawn from pedagogical examples. It originates from the publishing team's long-term research and consulting work on enterprise operations, industrial evolution, and public policy. We use 2022 multi-dimensional, industry-representative data as the foundation, organized into Minimal Business Logic Units, hereafter MBLUs, along an Enterprise, Industry, and Policy three-dimensional cross-analysis framework. During preprocessing, the team applies de-identification procedures that strip sensitive identifiers and potential model-knowledge-leakage risks while retaining the informational noise, missing values, and logical noise characteristic of real think-tank workflows. This design keeps the evaluation environment close to real business scenarios and avoids the inflated performance models can exhibit on idealized data.

\subsubsection{Task Extraction and Difficulty Stratification Strategy}

Based on the de-identified MBLUs, we extract high-frequency real-world problem sets from typical think-tank analytical workflows. Tasks are stratified into three difficulty tiers according to their cognitive complexity and cross-domain integration requirements.
\begin{itemize}
    \item \textbf{Easy} tasks focus on single-point fact retrieval, structured information extraction, and basic policy clause matching.
    \item \textbf{Medium} tasks require multi-source information alignment, causal inference, or policy-to-enterprise impact chain analysis.
    \item \textbf{Hard} tasks involve cross-domain data analysis, long-text comprehension, factual reasoning, or decision-making under complex constraints.
\end{itemize}

\subsubsection{Expert Recruitment and Annotation Guideline Development}

Annotation and quality verification are undertaken by an interdisciplinary expert team jointly established by a research institute and a university. Each candidate must have at least three years of research or graduate-level study experience in industrial economics, public policy, or corporate strategy, be proficient in quantitative and qualitative analytical methods, and pass benchmark assessments covering domain knowledge, logical reasoning, and text comprehension.

Prior to formal annotation, the technical team compiled the DataClawBench Annotation Guidelines. The guide underwent three rounds of pilot annotation and iterative refinement involving both human experts and AI agent testing. The final version covers task boundary definitions, quantitative criteria for difficulty classification, answer structure specifications, evidence traceability requirements, typical positive and negative samples, AI agent consensus thresholds and validation protocols, and contingency procedures for common ambiguities. The guidelines were finalized after pre-experimental validation of operability and calibration of multi-agent consistency standards, providing a standardized framework for subsequent human-AI collaborative annotation.

\subsubsection{Quality Control and Inter-Annotator Agreement}

To ensure the academic rigor and business applicability of the dataset, we designed a closed-loop quality control workflow comprising five stages of pilot annotation, consistency assessment, guideline iteration, formal annotation, and multi-level verification.
\begin{enumerate}
    \item For \textbf{Easy} difficulty, we use a pipeline combining domain knowledge graph construction, templated generation, and automated rule verification. After technical team review, 131 valid questions are retained.
    \item For \textbf{Medium} and \textbf{Hard} difficulty, an initial pool of over 400 high-value questions is manually curated by expert teams from the research institute and university. After the technical staff deliver the annotation guidelines, the university team performs the annotation. Each sample undergoes \textbf{back-to-back double-blind annotation} by at least two independent annotators, followed by cross-verification by third-party senior researchers who did not participate in the initial annotation.
    \item In \textbf{AI agent consensus verification}, each annotated sample is independently assessed by multiple AI agents against the annotation guidelines for rationality, evidentiary completeness, and domain validity. Only samples on which all agents reach unanimous agreement pass validation. Samples with divergent AI evaluations or failed validation are escalated to human experts for re-annotation. Controversial cases identified through this process are systematically analyzed and fed back into refining the guidelines.
    \item In \textbf{final verification}, the technical team conducts item-by-item review of AI-validated results against the annotation specifications, excluding samples with logical discontinuities, missing evidence, or deviations from business scenarios. The procedure yields 286 valid medium-difficulty and 75 valid hard-difficulty QA pairs.
\end{enumerate}

\section{Detailed Prompts}
\label{appendix:prompt}

This section reproduces the three prompt templates used by the DataClawBench evaluation pipeline. Placeholders in angle brackets are substituted at runtime from the task definition or from the agent's recorded trajectory.

\paragraph{Agent task prompt.}
Each agent under test receives a single user message assembled from the task's question, output guidelines, and a one-line statement of permitted data sources. Tasks in the International Comparison category additionally permit web search. All other categories restrict the agent to the local database mounted at \texttt{./database/}. The full template appears in the box below.

\begin{tcolorbox}[
  colback=gray!5,
  colframe=gray!40,
  title={Agent task prompt template},
  fonttitle=\bfseries\small,
  fontupper=\small,
  width=\linewidth,
  breakable
]
\begin{Verbatim}[
  fontsize=\small,
  breaklines=true,
  breaksymbolleft={},
  breakanywhere=true
]
<question>

Output guidelines:
<guidelines>

<data_sources>
\end{Verbatim}

The \texttt{<data\_sources>} line is set to either
``You may use files under \texttt{./database/} and web search.''
for International Comparison tasks, or
``Only use files under \texttt{./database/}.''
for all other categories.
\end{tcolorbox}

\paragraph{Outcome judge prompt for Acc.}
When the agent terminates, its final answer is graded by an LLM judge against the gold answer. The judge receives the original task prompt for context, an Expected Behavior summary, the agent's final text, and a category-dependent rubric. Single-answer tasks use a binary match rubric. Multi-part answer tasks use a per-part vector rubric whose total is the average of part scores.

\begin{tcolorbox}[colback=gray!5,colframe=gray!40,title={Outcome judge prompt template},fonttitle=\bfseries\small,width=\linewidth,fontupper=\small,breakable]
\begin{Verbatim}[fontsize=\small,
  breaklines=true,
  breaksymbolleft={},
  breakanywhere=true
]
You are a grading function. Your ONLY job is to output a single JSON object.

CRITICAL RULES:
- Do NOT use any tools (no Read, Write, exec, or any other tool calls)
- Do NOT create files or run commands
- Do NOT write any prose, explanation, or commentary outside the JSON
- Respond with ONLY a JSON object — nothing else

Be a strict evaluator. Judge the final assistant message against the task and rubric.

## Task
<task_prompt>

## Expected Behavior
<expected_behavior>

## Agent final answer
<agent_final_text>

## Grading Rubric
<rubric>

Score each criterion from 0.0 to 1.0.

Respond with ONLY this JSON structure (no markdown, no code fences, no extra text):
{"scores": {"criterion_name": 0.0}, "total": 0.0, "notes": "brief justification"}
\end{Verbatim}
\end{tcolorbox}

\paragraph{Process judge prompt for GPR.}
The process judge evaluates which gold-defined milestones were achieved during the agent's trajectory. It receives the gold reference steps, the milestone list, and a step-indexed reconstruction of the agent's execution. Tool outputs are pre-filtered to ``candidate'' snippets containing numeric matches against milestone values, reducing prompt size while preserving the evidence the judge needs. TPE is computed deterministically from the GPR result and per-milestone first-occurrence step indices, requiring no additional LLM call.

\begin{tcolorbox}[colback=gray!5,colframe=gray!40,title={Process judge prompt template for GPR},fonttitle=\bfseries\small,width=\linewidth,fontupper=\small,breakable]
\begin{Verbatim}[fontsize=\small,
  breaklines=true,
  breaksymbolleft={},
  breakanywhere=true
]
You are a milestone evaluator for an AI agent benchmark. Your task is to
determine which milestones an agent achieved during its execution trajectory.

## Reference Solution Path

The following ordered steps describe the gold-standard approach to solving this
task. Milestones are key intermediate results that should be produced along
this path.

### Gold Steps (in order)
<numbered gold steps>

### Milestones to Evaluate
<numbered M1..Mk milestones, each as "key: expected value">

## Agent Execution Trajectory

Below is a reconstruction of the agent's work, organized by step (assistant
turn). For each step, "Assistant" is the agent's own text output. "Candidate"
sections are excerpts from tool execution outputs where milestone-relevant
values were detected by numeric matching. These candidates may or may not be
true milestone achievements — you must verify the semantic context.

<per-step trajectory blocks>

## Evaluation Rules

1. Direct evidence: a milestone is achieved if the trajectory clearly shows
   the agent computed or obtained the expected value (or within 1% relative
   error) in the CORRECT semantic context. A number appearing in an unrelated
   context does NOT count.

2. Temporal coupling inference: milestones follow a logical dependency chain.
   If a downstream milestone is correctly achieved, its upstream dependencies
   can be inferred as achieved.

3. Chain-break identification: if the final answer is INCORRECT, identify the
   earliest milestone in the logical chain that was NOT achieved — this is
   the "break point" where the agent's reasoning diverged.

4. Different-but-valid paths: judge milestone achievement based on whether
   the agent obtained the correct intermediate values, regardless of method.

## Output Format

Respond with ONLY a JSON object. No markdown fences, no extra text.

{"milestones": [{"key": "...", "achieved": true, "evidence_type": "direct",
  "first_step": 4, "reason": "..."}, ...],
 "break_point": null,
 "chain_summary": "one-sentence summary"}
\end{Verbatim}
\end{tcolorbox}

\section{Case Study}
\label{appendix:casestudy}

We present four representative case studies that illustrate how process-oriented evaluation surfaces diagnostic signals invisible to outcome-only accuracy. The first three cases use Claude Opus 4.6, the strongest model on DataClawBench, and span a correct-but-inefficient trajectory, an incorrect trajectory with an early breakpoint, and an incorrect trajectory with a late breakpoint. The fourth case fixes one task and contrasts how five different models traverse it, surfacing the exploration archetypes from \S\ref{subsubsec:efficiency}. For readability, each trajectory below preserves the model's text reasoning verbatim where it is illuminating and summarizes tool calls together with their outputs. Gray boxes contain the task specification, and colored boxes contain the model trajectory.

\paragraph{Case 1. Correct answer, inefficient path.}
This task, drawn from the International Comparison category at medium difficulty, asks for ZEEKR's 2022 asset turnover ratio and its gap from the median of China's automobile manufacturing industry. The reference trajectory is four steps. Claude Opus 4.6 reaches the correct answer in 29 API requests, a 7.25$\times$ step expansion.

\begin{tcolorbox}[colback=gray!5,colframe=gray!40,title={Case 1 (Claude Opus 4.6, 29 requests, answer correct)},fonttitle=\bfseries\small,fontupper=\small,breakable]
\textbf{Question.} Against the backdrop of intensive policy rollouts for the new energy vehicle industry, what was ZEEKR's asset turnover ratio in 2022? How did it compare to the median asset turnover ratio of China's automobile manufacturing industry as a whole?

\textbf{Reference steps.}
\begin{enumerate}[itemsep=0pt,leftmargin=*,topsep=2pt]
\item Extract ZEEKR 2022 net revenues (RMB 31{,}899{,}448 thousand) and total assets (RMB 19{,}477{,}316 thousand) from the ZEEKR F-1 prospectus; compute asset turnover = 1.64.
\item From \texttt{company\_profile.csv}, filter enterprises with \texttt{industry = Automobile manufacturing}, yielding 230 enterprises.
\item From \texttt{company\_operation\_status.csv}, obtain revenue and total assets for the 230 enterprises; compute per-company asset turnover; compute industry median = 0.59.
\item ZEEKR minus industry median $= 1.64 - 0.59 = 1.05$.
\end{enumerate}
\textbf{Gold answer.} \texttt{[1.64, 1.05]}.

\vspace{0.7em}
\hrule
\vspace{0.7em}

\textbf{Claude Opus 4.6 trajectory.}
\begin{enumerate}[itemsep=2pt,leftmargin=*,topsep=2pt]
\item Opens with ``I'll start by checking what's available in the database and searching for the relevant financial data,'' and fires a directory listing plus a web search for ZEEKR in parallel.
\item Runs \texttt{head -5} against six CSV schemas simultaneously, an over-broad schema probe.
\item Greps for ZEEKR name variants across three profile files. All empty. Observes: ``The database uses anonymized company names.''
\item Pivots to an industry-based search. The keyword for automobile manufacturing yields 242 matches. Also scans the overseas-listing profile files in case ZEEKR was listed abroad.
\item Tries the parent company Geely. Finds only a supplier record that mentions ZEEKR as a customer, with no financials.
\item Fires a web search for the ZEEKR F-1 prospectus. SEC returns 403. First failed external fetch.
\item In parallel, pulls automobile-industry medians from \texttt{national\_industry\_status\_detail.csv}, obtaining median total assets and median operating revenue.
\item Recognizes that \texttt{median(revenue)/median(assets)} is not equivalent to \texttt{median(revenue/assets)} and commits to the per-company computation.
\item Runs a Python script over all 242 automobile-industry IDs to compute per-company ratios. First attempt returns an absurd median of 51{,}601.32 from a unit mismatch.
\item Diagnoses the bug: ``The units are NOT consistent -- different companies use different units.'' Values appear in yuan, 10K yuan, 100K yuan, 1M yuan, and 10M yuan.
\item Rewrites the script with a unit-conversion table, recomputes, and obtains industry median asset turnover = 0.59 across 230 companies.
\item Resumes the external hunt for ZEEKR's financials across additional sites. Two more web fetches fail or return garbled data.
\item Finally retrieves clean figures from a third-party analytics site. Total assets = RMB 19{,}477\,M, revenue = RMB 31{,}899\,M.
\item Computes $31{,}899 / 19{,}477 = 1.6378 \to 1.64$.
\item Runs a third Python verification that recomputes the industry median from scratch, then a fourth one-line recomputation of the final difference. Both are redundant.
\item Outputs \textbf{\texttt{[1.64, 1.05]}}.
\end{enumerate}

\textbf{Inefficiency pattern.} Each individual step is defensible, yet the compound trajectory exhibits five distinct sources of slack. First, schema probing before narrowing to the relevant tables. Second, double verification of ZEEKR's absence in the local database via multiple grep variants. Third, a unit-handling bug loop requiring two Python iterations. Fourth, four web-fetch dead-ends before a usable source is found. Fifth, two redundant re-verification passes on arithmetic that had already settled. Outcome-only evaluation records this task as a clean success, and EE captures the 7.25$\times$ inflation relative to the gold path.
\end{tcolorbox}

\paragraph{Case 2. Incorrect answer, early breakpoint on policy filtering.}
This task, drawn from the Comprehensive Decision category at hard difficulty, asks for the top province's weighted composite score in pharmaceutical manufacturing under a four-indicator scheme. The four indicators are enterprise agglomeration (weight 30\%), R\&D expenditure as a share of revenue (30\%), regional policy coverage intensity (20\%), and R\&D human resource penetration rate (20\%); policy coverage intensity is defined as each province's pharmaceutical-related policy count divided by the national total of pharmaceutical-related policies. The reference trajectory is eight milestones. Claude Opus 4.6 consumes 18 API requests and achieves 5 of 8 milestones, yet fails at M2 by counting all national policies (602) instead of the 80 pharmaceutical-related ones. The error propagates through the policy support indicator and inverts the final ranking, yielding 0.80 for Jiangsu instead of the gold answer 0.92 for Shanghai.

\begin{tcolorbox}[colback=gray!5,colframe=gray!40,title={Case 2 (Claude Opus 4.6, 18 requests, answer incorrect)},fonttitle=\bfseries\small,fontupper=\small,breakable]
\textbf{Question.} In 2022, a strategic consulting firm ranks provinces by their attractiveness in pharmaceutical manufacturing using a four-dimensional weighted scoring system. The four indicators, all min-max normalized then weighted, are enterprise agglomeration (30\%), R\&D expenditure as a share of revenue (30\%), regional policy coverage intensity (20\%), and R\&D human resource penetration rate (20\%). Policy coverage intensity is the ratio of relevant policy items in each province to the total number of relevant policies nationwide. What is the composite score of the top-ranked province?

\textbf{Reference steps.}
\begin{enumerate}[itemsep=0pt,leftmargin=*,topsep=2pt]
\item From \texttt{regional\_industry\_status.csv} filtered to pharmaceutical manufacturing, obtain the national enterprise total = 449.
\item From \texttt{policy\_release\_status.csv} filtered to pharmaceutical manufacturing, obtain the national pharmaceutical-related policy count = 80.
\item From \texttt{regional\_industry\_status.csv}, identify 16 provinces with complete pharmaceutical-manufacturing coverage.
\item Compute Shanghai industry agglomeration = 0.1203.
\item Compute Shanghai R\&D intensity = 0.2548.
\item Compute Shanghai policy support = 0.1375.
\item Compute Shanghai talent density = 0.162.
\item Compute Shanghai composite score after min-max normalization = 0.916.
\end{enumerate}
\textbf{Gold answer.} \texttt{0.92}.

\vspace{0.7em}
\hrule
\vspace{0.7em}

\textbf{Claude Opus 4.6 trajectory.}
\begin{enumerate}[itemsep=2pt,leftmargin=*,topsep=2pt]
\item Lists the database directory and probes the schemas of seven candidate CSVs in parallel, isolating \texttt{regional\_industry\_status.csv} and \texttt{policy\_release\_status.csv} as the relevant aggregate sources.
\item Filters \texttt{regional\_industry\_status.csv} to pharmaceutical manufacturing and aggregates per-province enterprise totals. Reaches 449 nationwide across 16 valid provinces, achieving M1 and M3.
\item \textbf{Break point.} Reads \texttt{policy\_release\_status.csv} and finds a row with \texttt{targetName = total published policy count, value = 602} and an empty \texttt{industryId}. Adopts 602 as the national denominator for ``relevant policies'' without filtering on \texttt{industry = pharmaceutical manufacturing}. The correct denominator, 80, sits in the same file under \texttt{industryId = 1} but is never queried.
\item For each province, divides the provincial pharmaceutical policy count by 602 to compute the policy intensity ratio, locking in a denominator that is 7.5 times too large.
\item Spends the next several requests debating the R\&D indicator. Catches an outlier-driven mean of 19960\% in one province, oscillates between mean and median, briefly attempts company-level aggregation across \texttt{company\_operation\_status.csv}, then returns to the regional pre-aggregated mean R\&D ratio. M5 and M7 both achieved with correct provincial values.
\item Recomputes the composite score with min-max normalization across provinces. Because the policy denominator is inflated, every province's policy contribution is uniformly compressed; provinces strong on policy support such as Shanghai lose ground and provinces strong on the other three axes such as Jiangsu move to the top.
\item Outputs \textbf{\texttt{0.80}} for Jiangsu, cross-checks with a mainland-only re-run that returns the same value, and confirms the answer. Incorrect.
\end{enumerate}

\textbf{Break-point analysis.} The failure is a single missed filter at M2 in the Policy Lookup and Count subtask category. Its structural cost is disproportionate to its locality. The flawed national denominator propagates linearly into M6 and rescales the relative weight of the four indicators after normalization, flipping the winning province from Shanghai to Jiangsu. Yet the agent still achieves 5 of 8 milestones, including every retrieval, unit-handling, and per-indicator computation step that does not depend on the policy denominator. Outcome-only evaluation collapses this run into a single 0.0 identical to a model that never reached any milestone, while the chain-break decomposition isolates the single false step that controls the answer.
\end{tcolorbox}

\paragraph{Case 3. Incorrect answer, late break point.}
This task, from the Enterprise--Industry Analysis category at medium difficulty, asks for the signed difference between the median operating profit of the Real Estate industry (containing company A) and that of the Financial Industry (containing company B). The reference trajectory is five steps. Claude Opus 4.6 achieves the first four milestones correctly but drops the sign on the final arithmetic, delivering an absolute value. Outcome-only scoring assigns 0.0, yet GPR shows the agent reaches 4/5 milestones before the break.

\begin{tcolorbox}[colback=gray!5,colframe=gray!40,title={Case 3 (Claude Opus 4.6, 14 requests, answer incorrect)},fonttitle=\bfseries\small,fontupper=\small,breakable]
\textbf{Question.} What is the difference between the median operating profit amount of the industry of Zhao Ye Hua Chang Real Estate Development Co., Ltd.\ and that of the industry of Tong Tong Ze Hong Securities Co., Ltd.?

\textbf{Reference steps.}
\begin{enumerate}[itemsep=0pt,leftmargin=*,topsep=2pt]
\item From \texttt{company\_profile.csv}, obtain the industry of Zhao Ye Hua Chang Real Estate Development Co., Ltd. = Real Estate.
\item From \texttt{national\_industry\_status.csv}, obtain the Real Estate median operating profit = 130{,}368{,}786 yuan.
\item From \texttt{company\_profile.csv}, obtain the industry of Tong Tong Ze Hong Securities Co., Ltd. = Financial Industry.
\item From \texttt{national\_industry\_status.csv}, obtain the Financial Industry median operating profit = 1{,}010{,}930{,}425 yuan.
\item Compute Real Estate minus Financial Industry $= 130{,}368{,}786 - 1{,}010{,}930{,}425 = -880{,}561{,}639.0$.
\end{enumerate}
\textbf{Gold answer.} \texttt{-880561639.0}.

\vspace{0.7em}
\hrule
\vspace{0.7em}

\textbf{Claude Opus 4.6 trajectory.}
\begin{enumerate}[itemsep=2pt,leftmargin=*,topsep=2pt]
\item Opens with ``Let me start by exploring the database structure and finding information about these two companies.'' Lists the database directory and previews the schema of \texttt{company\_profile.csv}.
\item Recognizes that the company names are pinyin transliterations of anonymized Chinese names and extracts the salient keywords Real Estate Development and Securities for substring matching.
\item Greps the Hua Chang Real Estate tag and finds five real-estate candidates. Only one carries the Zhao Ye prefix; Claude identifies it as the intended match.
\item Greps Tong Tong Ze Hong and locates the corresponding Securities firm. Both company-to-industry milestones cleanly achieved.
\item Records the mapping. Company A maps to Real Estate (industryId=4). Company B maps to Financial Industry (industryId=7).
\item Greps \texttt{national\_industry\_status.csv} for the median operating profit indicator in Real Estate. Retrieves 1303.68786 in units of 100{,}000 yuan.
\item Greps the same indicator for Financial Industry. Retrieves 101.0930425 in units of 10{,}000{,}000 yuan. Notes the mismatched units between the two rows.
\item Performs unit normalization. Real Estate: $1303.68786 \times 10^5 = 130{,}368{,}786$ yuan. Financial Industry: $101.0930425 \times 10^7 = 1{,}010{,}930{,}425$ yuan. Both conversions numerically correct.
\item \textbf{Break point.} Writes: ``The difference $= |1{,}010{,}930{,}425 - 130{,}368{,}786| = 880{,}561{,}639$ yuan.'' Silently wraps the subtraction in absolute-value bars and reorders the operands.
\item Briefly second-guesses the unit conversion rather than the sign, pivots to re-expressing everything in a common base unit, then stops without revisiting the arithmetic framing.
\item Outputs \textbf{\texttt{880561639.0}}. Incorrect, off by a sign.
\end{enumerate}

\textbf{Break-point analysis.} All four retrieval-and-normalization milestones are clean. The failure is a single absolute-value reflex applied to a signed quantity that the question explicitly defines as A minus B. Because the sign error occurs at the terminal milestone, outcome-only evaluation penalizes the task identically to Case~2, despite roughly 97\% of the reasoning chain being correct. The break-point decomposition, by contrast, records 4/5 milestones achieved and localizes the defect to ``difference'' interpretation, a diagnosable and distinct failure mode from early entity resolution.
\end{tcolorbox}

\paragraph{Case 4. The four archetypes on a single task.}
This task, drawn from the Industry Planning category at medium difficulty, asks which of two strategic routes scores higher for the province where the food-and-beverage enterprise with the most cumulative Chinese invention patent grants is located. Across the eight evaluated models, four reach the correct answer of industrial chain extension route and four fail; request counts span 4 to 83. We report five trajectories that cover all four archetypes from \S\ref{subsubsec:efficiency}.

\begin{tcolorbox}[colback=gray!5,colframe=gray!40,title={Case 4 (single task, five models compared)},fonttitle=\bfseries\small,fontupper=\small,breakable]
\textbf{Question.} In 2022, in the food and beverage industry, for the province where the enterprise with the most cumulative Chinese invention patent grants is located, if that province chooses the ``brand upgrade route'' (market-cap-to-revenue ratio, profit margin, per capita market cap; weights 35\%, 35\%, 30\%) versus the ``industrial chain extension route'' (total enterprises, revenue scale, upstream-downstream enterprise diversity; weights 40\%, 30\%, 30\%), which route scores higher?

\textbf{Reference steps.}
\begin{enumerate}[itemsep=0pt,leftmargin=*,topsep=2pt]
\item From \texttt{company\_operation\_status.csv}, identify the food-and-beverage enterprise with the most cumulative Chinese invention patent grants = Qingqing Jinyin Food Company, 644 patents.
\item From \texttt{company\_profile.csv}, obtain the company's province = Beijing.
\item From the regional aggregates and \texttt{company\_profile.csv}, compute Beijing's six per-route metrics across market-cap-to-revenue ratio, profit margin, per capita market cap, total enterprises, revenue scale, and upstream-downstream diversity, then apply the two weighted formulas after cross-province min-max normalization.
\item Brand upgrade route score = 25.0, industrial chain extension route score = 83.1.
\end{enumerate}
\textbf{Gold answer.} \texttt{Industrial chain extension route}.

\vspace{0.7em}
\hrule
\vspace{0.7em}

\textbf{Claude Opus 4.6 (20 requests, correct), the decisive solver.}
\begin{enumerate}[itemsep=2pt,leftmargin=*,topsep=2pt]
\item Lists the database directory, then probes four CSV schemas in parallel and isolates \texttt{Y\_EC\_44} as the cumulative-patent target field and \texttt{industryId=10} as the food-and-beverage industry.
\item Sorts the matching companies by patent count and lands on Qingqing Jinyin Food Company at 644 patents on the first sort, reads off the company's province as Beijing.
\item Pulls Beijing's six per-route metrics from the regional aggregates, normalizes across the provinces with complete data, and computes both route scores.
\item Returns the industrial chain extension route after one self-consistency check on the metric definitions.
\end{enumerate}

\textbf{Minimax M2.7 (42 requests, correct), persistent-but-late.}
\begin{enumerate}[itemsep=2pt,leftmargin=*,topsep=2pt]
\item Spends 41 silent assistant turns and roughly 60 tool calls scanning every profile and operation file for combinations of food, beverage, patent, and per-province aggregates before producing any user-facing text.
\item Surfaces Qingqing Jinyin Food Company in Beijing during the silent scan.
\item Computes the route scores directly from raw aggregate values, then re-checks ownership-type counts to reconstruct the upstream-downstream diversity metric.
\item At roughly twice Claude's request count, emits a single consolidated answer at the final turn that nominates the industrial chain extension route.
\end{enumerate}

\textbf{DeepSeek-V3.2 (83 requests, incorrect), wasteful trial-and-error.}
\begin{enumerate}[itemsep=2pt,leftmargin=*,topsep=2pt]
\item Locks onto Yili Weiwei Wine Company in Hubei at 324 patents as the patent leader, missing the higher-patent Qingqing Jinyin entry under the food sub-industry. M1 already broken.
\item Burns most of its remaining request budget trying to reconstruct Hubei's per-route metrics from company-level data with mismatched units, then switches to regional aggregates.
\item Cannot align targetName variants across provinces and concedes ``No relevant data found'' after burning the largest request budget on the task.
\end{enumerate}

\textbf{Qwen3.5-Plus (56 requests, incorrect), wasteful trial-and-error.}
\begin{enumerate}[itemsep=2pt,leftmargin=*,topsep=2pt]
\item Misreads the question's granularity, aggregating patent counts at the province level instead of selecting the individual top-patent enterprise.
\item Identifies Shanghai as the province with the highest aggregate patent count and attempts to compute the two routes for Shanghai.
\item Cannot recover an upstream-downstream diversity field, falls back to ``No relevant data found'' rather than reformulating the entity-selection step.
\end{enumerate}

\textbf{Kimi-K2.5 (4 requests, incorrect), disengaged.}
\begin{enumerate}[itemsep=2pt,leftmargin=*,topsep=2pt]
\item Lists the database directory and issues five schema-probe tool calls across the next two assistant turns, all silent and without committing to a candidate enterprise.
\item Returns a fourth assistant turn that emits neither tool calls nor text, leaving every gold milestone unachieved.
\end{enumerate}

\textbf{Style-contrast analysis.} The five trajectories trace the four archetypes from \S\ref{subsubsec:efficiency} on a single concrete task. Claude commits to one entity-selection hypothesis on its first sort and rides it to the answer in 20 requests. Minimax doubles the request count by exhaustively pre-scanning before committing and arrives at the same answer. DeepSeek and Qwen each burn fifty-plus requests on a wrong entity-selection branch, with DeepSeek picking the wrong individual enterprise (Hubei rather than Beijing) and Qwen picking the wrong abstraction level (provincial aggregate rather than top enterprise). Kimi terminates after four requests without ever attempting analysis. The correct-versus-incorrect split on this task is an entity-selection gap rather than a reasoning-depth gap. Every model that locked onto the right enterprise on its first or second attempt produced a correct answer, and every model that committed to a wrong entity propagated the error through dozens of request cycles or skipped the work entirely.
\end{tcolorbox}

\end{document}